\documentclass{article} 
\let\standardaddcontentsline\addcontentsline
\usepackage{iclr2027_conference,times}
\let\iclraddcontentsline\addcontentsline
\let\addcontentsline\standardaddcontentsline

\usepackage{amsmath,amsfonts,bm}

\def\eqref#1{equation~\ref{#1}}

\def\1{\bm{1}}

\DeclareMathAlphabet{\mathsfit}{\encodingdefault}{\sfdefault}{m}{sl}
\SetMathAlphabet{\mathsfit}{bold}{\encodingdefault}{\sfdefault}{bx}{n}

\usepackage{xcolor}
\definecolor{mydarkred}{rgb}{0.6,0,0}
\definecolor{myblue}{HTML}{268BD2}
\usepackage[
colorlinks,
linkcolor=mydarkred,
citecolor=myblue
]{hyperref}
\usepackage{graphicx}
\usepackage{wrapfig}
\usepackage{float}
\usepackage[most]{tcolorbox}
\usepackage{enumitem}
\usepackage{url}

\newcommand{\method}{CARing}
\title{Coverage-Aware Reasoning with Medical Tokens for Diagnosis Prediction}

\author{
\textbf{Kaisong Zhang$^{1}$\thanks{Equal contribution.}\hspace{0.5em}, Haotian Fang$^{2}$\footnotemark[1]\hspace{0.5em}, Junmeng Zhou$^{3}$, Hang Lv$^{1}$, Yulan Pan$^{1}$, Yanchao Tan$^{1}$\thanks{Corresponding author.}} \\
\textnormal{$^1$ Fuzhou University \quad $^2$ Shanghai Jiao Tong University \quad $^3$ Beihang University} \\
}

\iclrfinalcopy 
\begin{document}
\let\appendixaddcontentsline\addcontentsline
\let\addcontentsline\iclraddcontentsline

\maketitle

\begin{abstract}
    Large language models (LLMs) offer promising potential for next-visit
    diagnosis prediction, owing to their ability to integrate longitudinal
    clinical evidence and reason over it in natural language. 
    However, reinforcement learning for LLM reasoning commonly rewards each
    trajectory according to the correctness of its final answer.
    In next-visit diagnosis prediction, multiple diagnoses can be simultaneously
    valid, but independently rewarding one diagnosis per trajectory does not
    distinguish repeated hits from coverage of different diagnoses.
    The policy can therefore concentrate on a few correct diagnoses,
    leaving others uncovered.
    Meanwhile, LLM tokenizers can split ICD codes into several generic tokens with limited clinical meaning, requiring multiple decoding steps to predict each diagnosis and hindering reasoning over a
    large disease vocabulary. To address both challenges, we propose \method{}, 
    a framework that represents diagnoses with compositional Semantic IDs (SIDs) and optimizes 
reasoning trajectories for multi-label coverage.
    Concretely, we first encode ontology-enriched disease semantics into compact SIDs
    through residual quantization, and ground the resulting SID tokens in
    natural language and longitudinal EHR contexts through multi-task alignment and
    reasoning-enriched training to unlock transferable LLM reasoning. \method{} further
    improves unordered multi-label prediction through a coverage reward for
    reinforcement learning and multi-positive supervision.
    At inference time, the
    model supports both efficient direct constrained decoding and multi-chain
    reasoning with rank fusion. On MIMIC-III and MIMIC-IV, \method{} exceeds all EHR-trained baselines in weighted $F_1$ and attains the highest top-$k$ recall at every reported cutoff, including R@30 of 46.04\% and 46.52\% in reasoning mode.
    Our codes and logs are available at {\urlstyle{same}\url{https://github.com/zmlxzyh/CARing-Codes-Logs}}.
    
\end{abstract}

\vspace{-2pt}
\section{Introduction}
\begin{figure}[t]
    \centering
    \includegraphics[width=\textwidth]{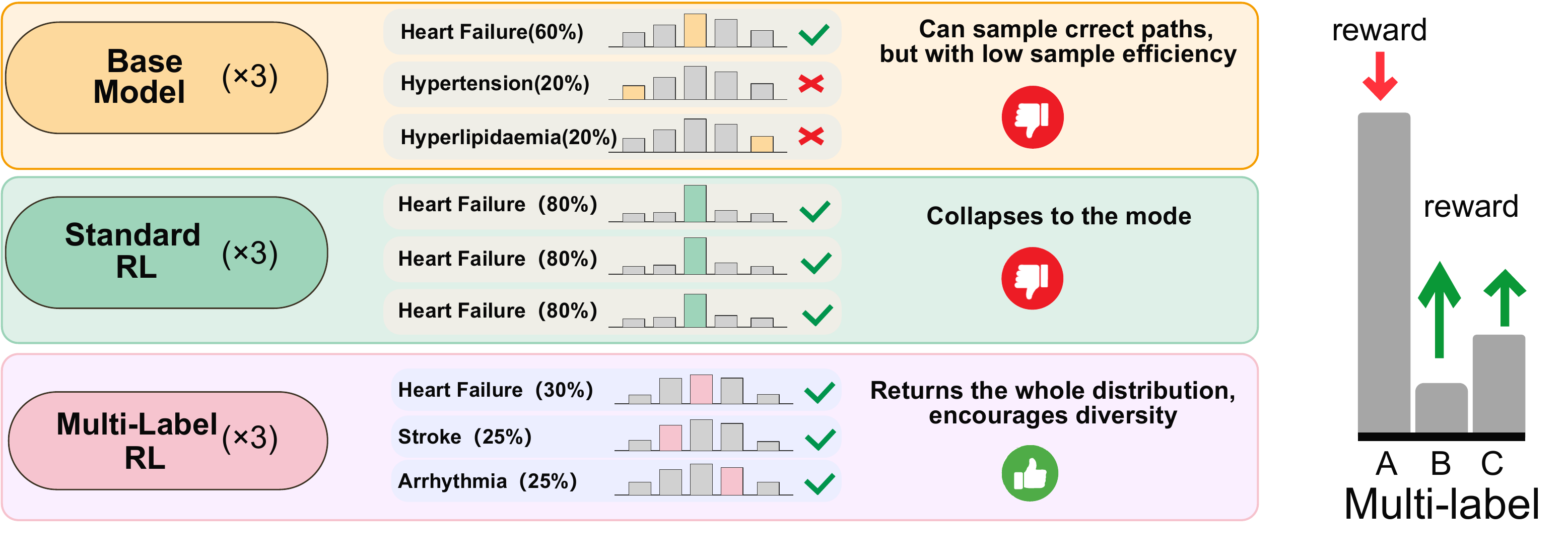}
    \vspace{-1em}
    \caption{Motivation for multi-label reinforcement learning in next-visit
diagnosis prediction. Standard RL can collapse onto one correct diagnosis,
leaving others uncovered. Our coverage reward discounts repeated predictions
and encourages coverage of distinct correct diagnoses.}
    \label{fig:multilabel-motivation}
    \vspace{-0.5em} 
\end{figure}

\vspace{-2pt}
Next-visit diagnosis prediction aims to infer diagnoses at a patient's
next encounter from their previous visits.
Existing EHR predictors learn from longitudinal patient histories
\citep{choi2016doctor,ma2018kame,rasmy2021med,chen2024predictive,KE2026104087}.
Large language models (LLMs) offer a way to incorporate explicit reasoning
into this task by synthesizing historical evidence and generating
intermediate clinical hypotheses in natural language
\citep{jiang2025reasoning,cao2026remedi}.
Reinforcement learning is commonly used to improve LLM reasoning by
rewarding the correctness of each trajectory's final answer
\citep{shao2024deepseekmath,liu2025beyond}.
However, next-visit diagnosis prediction is an
\emph{unordered multi-label prediction} problem over a large diagnosis
vocabulary. Multiple diagnoses can be simultaneously correct, and their
order within a visit has no intrinsic meaning.
When each trajectory predicts one diagnosis and is rewarded only for its
own correctness, a group that repeats one correct diagnosis earns the same
total reward as an equally sized group covering different correct
diagnoses. The policy can therefore concentrate on a few correct
diagnoses and leave others uncovered \citep{yue2025does,gxchen2025kl},
as illustrated in Figure~\ref{fig:multilabel-motivation}.

Effective reasoning also requires a diagnosis representation that an LLM can
understand and generate efficiently. Although ICD codes organize diagnoses
hierarchically, their letters and numbers do not describe the diseases in
ordinary language. LLM tokenizers can split a single code
into several character-level tokens rather than treat it as a disease-level
unit, and such fragmented codes are hard to ground, reason over, and generate
across a large diagnosis vocabulary~\citep{soroush2024large}. Medical
tokenizers such as MedTok and MedRQ compress code descriptions, relations,
and ontologies into reusable discrete codebooks
\citep{su2025multimodal,lv2026towards}, but their downstream predictors use
the resulting representations as model inputs rather than as tokens for
reasoning and prediction. MERA maps each disease to an atomic special token
and handles the unordered diagnoses within a visit~\citep{ma2025memorize}, but
its vocabulary and associated parameters grow linearly with the number of
diagnoses, and natural language reasoning stays outside its prediction path.
Semantic IDs (SIDs) in recommendation suggest a compositional alternative:
items reuse codebook tokens~\citep{rajput2023recommender,zheng2024adapting},
and recent systems let an LLM reason in natural language before generating an
identifier~\citep{liu2025onerec,team2026onereason}. Inspired by this approach,
we use medical SIDs as a shared token language linking disease concepts to
longitudinal patient histories. Unlike recommendation, where the target is
usually a single item, a future visit can contain multiple correct diagnoses,
so the model must recover complementary labels.



We propose \method{} (\textbf{C}overage-\textbf{A}ware
\textbf{R}easoning with Compositional Medical Tokens for
Multi-Label Next-Visit Diagnosis Prediction), a framework that brings LLM
reasoning to diagnosis prediction through medical SIDs and trains reasoning
trajectories to cover multiple correct diagnoses.

Our technical contributions are as follows:
\begin{itemize}[leftmargin=*]
    \item \textbf{Adapting LLM reasoning to diagnosis prediction through medical SIDs.}
    We use medical SIDs to connect clinical language, patient histories,
    and diagnosis outputs within one autoregressive model.
    Disease semantics and ICD ontology paths are compressed through
    residual quantization into shared codebook tokens.
    Six alignment tasks and narratives generated by a teacher LLM
    ground these tokens in clinical context.
    Reasoning activation then trains the model to generate a clinical
    rationale before predicting a diagnosis SID.

    \item \textbf{Optimizing reasoning trajectories for diagnosis coverage.}
    A coverage reward divides each correct prediction's credit by its
    frequency within a group of sampled trajectories, so the total group
    reward equals the number of distinct correct diagnoses.
    An auxiliary multi-positive objective supervises only the SID answer
    tokens, supplying learning signals for correct diagnoses that policy
    sampling may miss. These objectives encourage trajectories to recover complementary diagnoses.

    \item \textbf{Direct prediction and explicit reasoning within a common inference framework.}
    The direct mode ranks diagnoses through constrained decoding without
    generating a rationale. The reasoning mode samples multiple chains of
    thought before prediction, ranks diagnoses by constrained decoding after each chain, and
    combines the rankings with reciprocal rank fusion.
    Both modes restrict predictions to valid medical SIDs.
\end{itemize}
\vspace{-3pt}

We evaluate \method{} on MIMIC-III and MIMIC-IV
\citep{johnson2016mimic,johnson2023mimic}.
\method{} outperforms all baselines trained on EHR data in weighted
$F_1$ and achieves the highest top-$k$ recall among all evaluated
methods.
The reasoning mode reaches R@30 of 46.04\% and 46.52\% on the two
datasets, respectively.
Ablation studies show that semantic enrichment improves direct prediction
and supports the complementary roles of the coverage reward and
multi-positive supervision.
\vspace{-2pt}
\section{Related Work}
\label{sec:related-work}
\vspace{-2pt}
We briefly summarize related work and refer readers to
Appendix~\ref{appendix:related-work} for a detailed discussion.

Next-visit diagnosis prediction is commonly addressed by EHR models that encode
visit sequences with recurrent or attention architectures
\citep{choi2016doctor,choi2016retain,gao2020stagenet,yang2023transformehr} and
incorporate medical ontologies or patient graphs
\citep{ma2018kame,lu2021collaborative,chen2024predictive,tan2025boxlm}. These
models output scores over a fixed diagnosis vocabulary rather than generating
diagnoses through explicit reasoning. Bringing LLMs to this task requires
diagnosis tokens that the model can read and generate. Medical tokenizers learn
discrete codes for diseases but mainly feed them to downstream predictors
\citep{su2025multimodal,lv2026towards}, while MERA adds one vocabulary token per
disease, so the vocabulary grows with the number of diagnoses
\citep{ma2025memorize}. Semantic IDs in recommendation instead compose each item
from shared codebook tokens, and recent systems let an LLM reason in text before
generating the identifier \citep{liu2025onerec,team2026onereason}. Explicit
reasoning has also improved clinical prediction
\citep{jiang2025reasoning,cao2026remedi}, but mainly for binary outcomes or small
label spaces. Our work represents diseases with compositional medical SIDs, so
that one LLM reads patient histories, reasons in natural language, and generates
diagnoses from a large vocabulary in which a visit has several correct labels.

Large reasoning models generate intermediate steps before the final answer
\citep{wei2022chain} and gain accuracy from sampling multiple chains at test time
\citep{wang2023self,snell2024scaling}. Reinforcement learning with verifiable
rewards (RLVR) trains such reasoning from outcome feedback
\citep{shao2024deepseekmath,guo2025deepseek}, but rewarding each trajectory only
for its own correctness tends to concentrate the policy on a few answers. RLVR
raises pass@1 while base models reach higher pass@$k$ at large $k$
\citep{yue2025does}, and weak KL regularization with equal rewards tends to
specify a unimodal target distribution \citep{gxchen2025kl}. Recent methods
jointly optimize diversity and correctness by scoring the whole rollout group \citep{anschel2025group, li2026setpo}.
In diagnosis prediction, recall depends on how many correct diagnoses of a visit
the model covers. Our coverage reward divides each correct SID's credit by its
frequency in the group, so the group reward counts distinct correct diagnoses,
and a multi-positive loss supplies signal for correct diagnoses that no
trajectory samples.

\vspace{-2pt}
\section{Method}
\label{sec:method}
\begin{figure}[t]
    \centering
    \includegraphics[width=\textwidth]{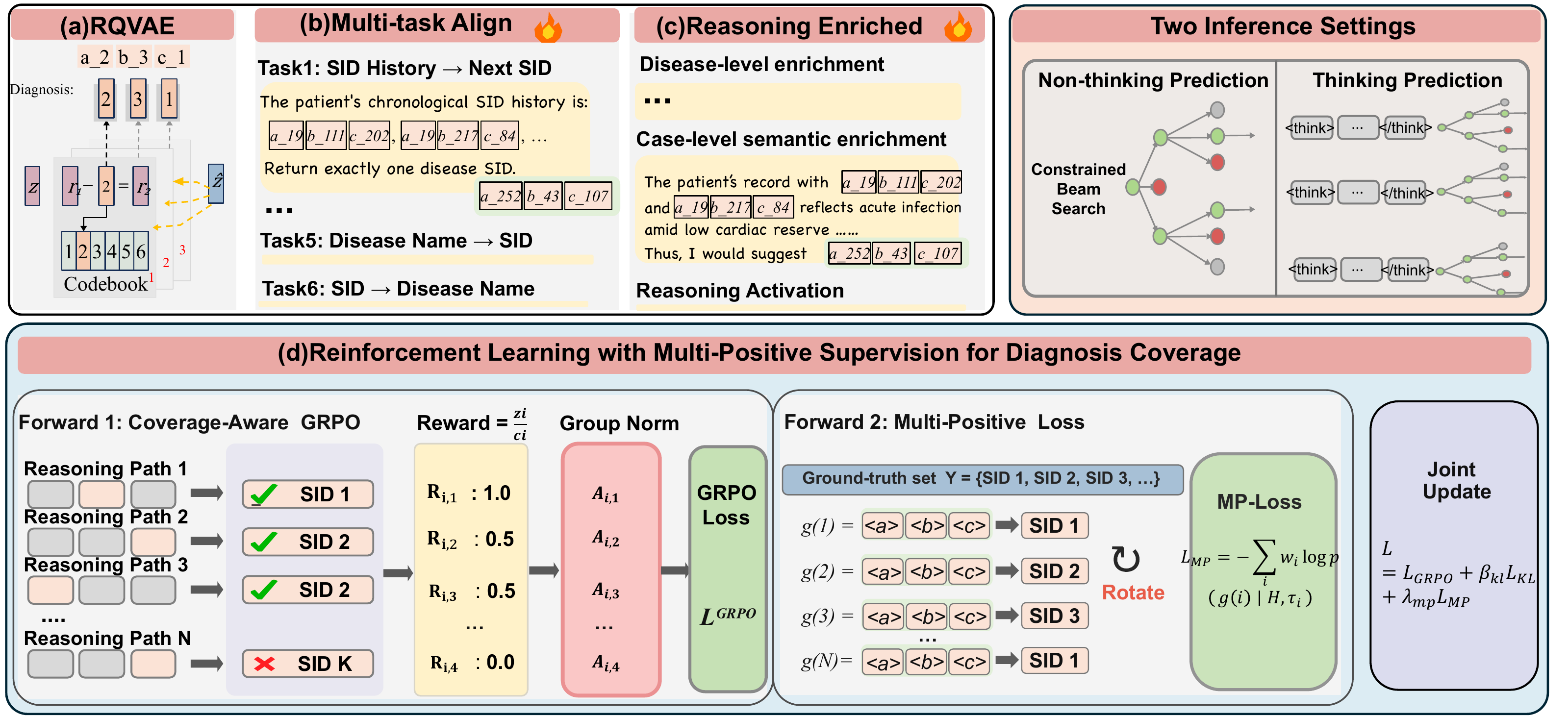}
    \vspace{-1em}
    \caption{Overview of \method{}. The framework constructs compositional medical SIDs, aligns them with clinical language and reasoning, combines a coverage reward with multi-positive answer supervision, and supports direct and multi-chain reasoning inference.}
    \label{fig:framework}
    \vspace{-0.5em}
\end{figure}

\vspace{-2pt}
Our framework first constructs ontology-enriched medical SIDs from shared tokens, then grounds these tokens in clinical text and longitudinal patient histories, and finally optimizes reasoning trajectories for multi-label diagnosis coverage. Figure~\ref{fig:framework} illustrates the overall pipeline.

\vspace{-3pt}
\subsection{Task Formulation}
\label{sec:task-formulation}
\vspace{-3pt}

Let $\mathcal{D}$ denote the diagnosis vocabulary. A patient's longitudinal
record up to visit $t$ is $H_t=(V_1,\ldots,V_t)$, where
$V_j\subseteq\mathcal{D}$ is the set of diagnoses at visit $j$.
The prediction target is the unordered diagnosis set at the next visit,
$Y_t=V_{t+1}=\{d_1,\ldots,d_m\}\subseteq\mathcal{D}$.

To avoid imposing an order on the target diagnoses, we expand each
visit-level example $(H_t,Y_t)$ into case-level instances
$\{(H_t,d_i)\}_{i=1}^{m}$. Each instance pairs the same history with
one target. We use them for SID--language alignment
and reasoning activation, with each trajectory predicting one diagnosis
SID. Reinforcement learning and visit-level evaluation retain the
complete target set $Y_t$.
\vspace{-3pt}
\subsection{Ontology-Enriched Compositional Medical Tokenization}
\label{sec:medical-tokenization}
\vspace{-3pt}
\paragraph{Ontology-enriched disease semantics.}
For each disease $d\in\mathcal{D}$, we pair its canonical title $n(d)$ with its root-to-leaf ICD ontology path $p(d)$, represented schematically as $u_d=\texttt{Disease: }n(d)\ \oplus\ \texttt{ICD ontology path: }p(d)$. We encode the textual metadata $u_d$ into a continuous semantic representation $\mathbf{x}_d$, which is then fed into a residual quantization module.

\paragraph{Residual quantization.}
Following residual-quantized VAEs (RQ-VAEs)~\citep{zeghidour2021soundstream}, an MLP encoder maps $\mathbf{x}_d$ to a latent vector $\mathbf{z}_d=E(\mathbf{x}_d)\in\mathbb{R}^{d_z}$. Given $L$ codebooks $\mathcal{C}_l=\{\mathbf{e}^{(l)}_k\}_{k=1}^{K}$, residual quantization starts with $\mathbf{r}^{(0)}_d=\mathbf{z}_d$. At each level, it selects $c_d^{(l)}=\arg\min_k\|\mathbf{r}^{(l)}_d-\mathbf{e}^{(l)}_k\|_2^2$ and updates the residual as $\mathbf{r}^{(l+1)}_d=\mathbf{r}^{(l)}_d-\mathbf{e}^{(l)}_{c_d^{(l)}}$. The quantized latent $\mathbf{z}_{q,d}=\sum_{l=0}^{L-1}\mathbf{e}^{(l)}_{c_d^{(l)}}$ is decoded back to the original embedding space. The entry selected at level $l=0,\ldots,L-1$ becomes medical token $s_d^{l+1}$; these tokens form the SID sequence $q(d)=(s_d^1,\ldots,s_d^L)$, with $L=3$ in our experiments. These components are jointly trained with $\mathcal{L}_{\mathrm{tok}}=\|\mathbf{x}_d-D(\mathbf{z}_{q,d})\|_2^2+\frac{1}{L}\sum_{l=0}^{L-1}\bigl(\|\mathrm{sg}[\mathbf{r}^{(l)}_d]-\mathbf{e}^{(l)}_{c_d^{(l)}}\|_2^2+\beta\|\mathbf{r}^{(l)}_d-\mathrm{sg}[\mathbf{e}^{(l)}_{c_d^{(l)}}]\|_2^2\bigr)$, where $\mathrm{sg}[\cdot]$ is stop-gradient and $\beta$ controls the commitment term. The resulting SID is a compact sequence of discrete tokens that preserves disease semantics for autoregressive modeling. We write $H$ for the SID-encoded history $H_t$, preserving visit boundaries.
\vspace{-3pt}
\subsection{Multi-Task SID--Clinical Context Alignment}
\label{sec:multitask-alignment}
\vspace{-3pt}

Although the RQ-VAE preserves disease-level similarity, its component tokens are newly added to the LLM vocabulary and initially have no lexical meaning. To enable the LLM to interpret, reason over, and generate SID tokens in clinical contexts, we conduct full-parameter autoregressive training on six complementary tasks, where SID tokens and natural language tokens co-occur across diverse medical scenarios. Four \emph{next-visit prediction} tasks cover the combinations required by the downstream model: SID history $\rightarrow$ SID (T1), textual disease history $\rightarrow$ SID (T2), SID history $\rightarrow$ disease name (T3), and textual disease history $\rightarrow$ disease name (T4). Two \emph{concept translation} tasks establish bidirectional grounding: disease name $\rightarrow$ SID (T5) and SID $\rightarrow$ disease name (T6). T1--T4 jointly align textual and SID histories with both language and SID outputs, with T1 matching the final prediction interface; T5--T6 teach the SID vocabulary to encode recognizable disease semantics. All tasks are jointly optimized under the standard next-token prediction objective.
Due to space limitations, detailed input and output formats for all six tasks are provided in Appendix~\ref{appendix:multitask-task-formats}.



\vspace{-3pt}
\subsection{Reasoning-Enriched Training}
\label{sec:reasoning-enriched-training}
\vspace{-3pt}
The six alignment tasks ground SIDs through concise translation and prediction templates, but expose them to limited linguistic and clinical contexts. To address this issue, we leverage a large teacher language model with rich world knowledge. We augment SID--language alignment with two offline teacher-generated corpora and general-domain reasoning data, then train the aligned model to follow an explicit reason-then-predict pathway.

\paragraph{Disease-level semantic enrichment.}
For each disease, the teacher generates a clinical narrative in which its SID denotes the concept across manifestations, investigations, comorbidities, and ontology relations. To ground the description, each prompt includes metadata from the PyHealth ICD-9-CM ontology~\citep{yang2023pyhealth} and the 2015 Clinical Classifications Software (CCS) and Chronic Condition Indicator (CCI) released by the AHRQ Healthcare Cost and Utilization Project~\citep{ahrq2015ccs,ahrq2015cci}. The resulting text interleaves SID tokens with natural language. Training on the entire narrative with next-token prediction exposes each SID to clinical language beyond the fixed T1--T6 templates.

\paragraph{Case-level semantic enrichment.}
For each case-level training instance, the teacher first receives the historical SID visits and a reference target used only as private guidance, and generates a reasoning path $r$. A second teacher call integrates the history, $r$, and the target into a coherent narrative. 

We concatenate these two enriched corpora, the T1--T6 datasets, and general-domain reasoning data for continued full-parameter alignment training. The general data are intended to mitigate degradation of the backbone's language and reasoning abilities.

\paragraph{Reasoning activation.}
Rich alignment does not by itself ensure that the model will externalize reasoning before predicting a SID. We reuse the teacher-generated reasoning path $r$ from case-level semantic enrichment to construct supervised samples for each history--diagnosis pair $(H,d)$. Starting from the enriched alignment checkpoint, we apply supervised fine-tuning to history-only prompts, supervising $r$ followed by the target SID $q(d)$ while masking the prompt tokens. This fine-tuning targets instruction following and the reason-then-SID response format.

Representative training sequences are provided in Appendix~\ref{appendix:reasoning-enriched-task-formats}, and their objectives are defined in Appendix~\ref{appendix:reasoning-enriched-objectives}. The teacher prompts for data enrichment are provided in Appendix~\ref{appendix:reasoning-enriched-data-prompts}.

\vspace{-3pt}
\subsection{Reinforcement Learning with Multi-Positive Supervision}
\label{sec:accuracy and diversity-aware-rl}
\vspace{-3pt}

Reasoning activation uses case-level instances with one target SID, whereas a future visit generally contains multiple unordered correct diagnoses. An exact match (EM) reward assigns the same credit to every correct prediction, even when several trajectories repeat one diagnosis. It therefore cannot distinguish repeated hits from coverage of different diagnoses. We address this mismatch with GRPO~\citep{shao2024deepseekmath} at the visit level, combining a coverage reward over sampled trajectories with supervision over the full positive set.

\paragraph{Sampling by visit.}
For one history $H$ with ground-truth SID set $Y=\{q(d):d\in V_{t+1}\}$, the old policy samples a group of $N$ reasoning--prediction trajectories
\begin{equation}
    o_i=(\tau_i,y_i)\sim\pi_{\theta_{\mathrm{old}}}(\cdot\mid H),
    \qquad i=1,\ldots,N,
    \label{eq:visit-rollouts}
\end{equation}
where $\tau_i$ is the generated reasoning and $y_i$ is the final three-token SID. Let $a_{i,1},\ldots,a_{i,T_i}$ be the tokens of serialized $o_i$ included in the loss.


\paragraph{Coverage reward.}
Let $z_i=\mathbf{1}[y_i\in Y]$ indicate a correct prediction and let $c_i=\sum_{j=1}^{N}\mathbf{1}[y_j=y_i]$ be its frequency in the rollout group. We assign
\begin{equation}
    r_i=\frac{z_i}{c_i}.
    \label{eq:coverage-reward}
\end{equation}
Each distinct correct SID predicted by the group contributes one unit to the total reward, regardless of how often it appears (Appendix~\ref{appendix:coverage-reward-property}). When rewards vary within the group, GRPO assigns larger advantages to less-repeated correct SIDs than to repeated hits, encouraging trajectories to cover complementary correct diagnoses.

\paragraph{Policy optimization.}
After computing the shaped rewards, GRPO compares trajectories generated from the same history without training a value model. Let $\mu_H$ and $\sigma_H$ be the mean and standard deviation of the $N$ rewards. Each rollout receives the normalized outcome advantage
\begin{equation}
    \mu_H=\frac{1}{N}\sum_{j=1}^{N}r_j,\qquad
    \sigma_H=\operatorname{Std}(\{r_j\}_{j=1}^{N}),\qquad
    \widehat{A}_i=\frac{r_i-\mu_H}{\sigma_H+\epsilon}.
    \label{eq:grpo-advantage}
\end{equation}
Here $\epsilon>0$ stabilizes the normalization.
For token $t$, $h_{i,t}=(H,a_{i,<t})$ is its context and $\rho_{i,t}(\theta)=\pi_\theta(a_{i,t}\mid h_{i,t})/\pi_{\theta_{\mathrm{old}}}(a_{i,t}\mid h_{i,t})$ is its importance ratio; $\varepsilon>0$ is the clipping width.
\begin{equation}
    \mathcal{L}_{\mathrm{GRPO}}
    =-\frac{1}{\sum_i T_i}\sum_{i=1}^{N}\sum_{t=1}^{T_i}
    \min\!\left[
        \rho_{i,t}(\theta)\widehat{A}_i,\,
        \operatorname{clip}(\rho_{i,t}(\theta),1-\varepsilon,1+\varepsilon)
        \widehat{A}_i
    \right].
    \label{eq:grpo-policy-loss}
\end{equation}
The same $\widehat{A}_i$ applies to the reasoning and SID answer tokens. Minimizing this objective increases the likelihood of reasoning and answer tokens in above-average trajectories and decreases it for below-average trajectories. Further optimization details are given in Appendix~\ref{appendix:grpo-optimization}.

\paragraph{Multi-positive supervision.}
Coverage reward gives no policy-gradient signal for correct diagnoses absent
from the rollout group. We therefore supervise alternative positives at the
answer level. Let $\mathcal I$ index rollouts with an identifiable three-token
terminal SID. For $i\in\mathcal I$, we retain its reasoning prefix $\tau_i$,
replace the answer with $g(i)\in Y$, and optimize its $L$ SID tokens,
writing $g(i)_l$ for token $l$ and $g(i)_{<l}$ for its prefix:
\begin{equation}
    \mathcal{L}_{\mathrm{MP}}
    =-\frac{1}{N}\sum_{i\in\mathcal I}w_i\,
    \frac{1}{L}\sum_{l=1}^{L}
    \log\pi_{\theta}\!\left(
        g(i)_l\mid H,\tau_i,g(i)_{<l}
    \right).
    \label{eq:multi-positive-loss}
\end{equation}
We assign $g(i)$ round-robin over a fixed enumeration of $Y$, with a
step-dependent starting offset. For $S=\{g(i):i\in\mathcal I\}$ and
$n_g=|\{i\in\mathcal I:g(i)=g\}|$, we set $w_i=N/(|S|n_{g(i)})$, giving each
selected SID equal total weight. The loss is zero when $\mathcal I$ is empty.

\paragraph{Joint optimization.}
The actor is optimized with
\begin{equation}
    \mathcal{L}
    =\mathcal{L}_{\mathrm{GRPO}}
    +\beta_{\mathrm{KL}}\,\widehat{D}_{\mathrm{KL}}
        (\pi_{\theta}\Vert\pi_{\mathrm{ref}})
    +\lambda_{\mathrm{MP}}\mathcal{L}_{\mathrm{MP}},
    \label{eq:joint-rl-objective}
\end{equation}
where $\beta_{\mathrm{KL}}$ and $\lambda_{\mathrm{MP}}$ weight the KL and
multi-positive terms, and $\pi_{\mathrm{ref}}$ is the fixed reasoning activation
policy; see Appendix~\ref{appendix:grpo-optimization} for gradients.
\vspace{-3pt}
\subsection{Two Inference Settings}
\label{sec:inference-settings}
\vspace{-3pt}
Because next-visit diagnoses form an unordered set, we rank individual diagnosis SIDs rather than generate the full set as one autoregressive sequence. Both settings use catalog-constrained beam search to obtain valid, deduplicated candidates (Appendix~\ref{appendix:constrained-beam-search}).

\paragraph{Direct inference (Non-Thinking).}
Multi-task alignment and reasoning-enriched training strengthen the checkpoint's direct prediction capability, enabling efficient inference without explicitly generating a reasoning chain. For efficient prediction, we disable explicit thinking and condition the model only on the longitudinal SID history. Constrained beam search directly expands valid SID prefixes and ranks the completed paths by their autoregressive sequence scores,
\begin{equation}
    S_{\mathrm{dir}}(d\mid H)
    =\sum_{l=1}^{L}
    \log p_{\theta}\!\left(s_d^l\mid H,s_d^{<l}\right).
    \label{eq:direct-inference}
\end{equation}
Duplicate SID paths are removed before returning the ranked list. 

\paragraph{Multi-chain reasoning (Thinking).}
We use the checkpoint after joint optimization with the coverage reward and multi-positive loss. From the same history, we sample $R$ reasoning chains $\{\tau_r\}_{r=1}^{R}$. For each chain, constrained decoding yields a SID ranking that is mapped to a diagnosis ranking $\mathcal{R}_r$. We rank the union of these candidates using reciprocal rank fusion (RRF)~\citep{cormack2009reciprocal}:
\begin{equation}
    S_{\mathrm{RRF}}(d\mid H)
    =\sum_{r:\,d\in\mathcal{R}_r}
    \frac{1}{\kappa+\operatorname{rank}_{\mathcal{R}_r}(d)},
    \label{eq:rrf}
\end{equation}
where $\kappa>0$ controls the weight assigned to top-ranked candidates. 

\vspace{-2pt}
\section{Experiments}
\label{sec:experiments}
\vspace{-2pt}

In this section, we aim to answer the following research questions:

\begin{itemize}[leftmargin=*,itemsep=1pt,topsep=2pt,parsep=0pt]
    \item \textbf{RQ1:} How does \method{} compare with ML and LLM baselines?
    \item \textbf{RQ2:} Do ICD ontology paths improve SID quality and diagnosis recall?
    \item \textbf{RQ3:} How do different design choices affect the performance of \method?
    \item \textbf{RQ4:} How do the coverage reward and multi-positive loss affect diagnosis coverage and diversity?

\end{itemize}

\paragraph{Datasets.}
We conduct experiments on two real-world electronic health record datasets, MIMIC-III~\citep{johnson2016mimic} and MIMIC-IV~\citep{johnson2023mimic}, following the preprocessing protocol of prior diagnosis-prediction studies~\citep{ma2018kame,chen2024predictive}. Due to the larger scale of MIMIC-IV, we sample 10,000 patients for our experiments. Detailed preprocessing procedures and dataset statistics are provided in Appendix~\ref{appendix:dataset-preprocessing}.
\paragraph{Evaluation Metrics.}
We adopt weighted $F_1$ score (w-$F_1$)~\citep{bai2018interpretable,lu2021collaborative} and top-$k$ recall (R@$k$)~\citep{choi2016doctor,lu2021collaborative} for next-visit diagnosis prediction. w-$F_1$ is a weighted sum of the $F_1$ score for each diagnosis. R@$k$ is the ratio of ground-truth diagnoses retrieved among the top-$k$ predictions to the total number of ground-truth diagnoses. 

\paragraph{Baselines.}
We compare with four groups of baselines: (1) \emph{neural EHR predictors}, including RETAIN~\citep{choi2016retain}, KAME~\citep{ma2018kame}, StageNet~\citep{gao2020stagenet}, CGL~\citep{lu2021collaborative}, and TRANS~\citep{chen2024predictive}; (2) a \emph{PLM-based predictor}, BoxLM~\citep{tan2025boxlm}; (3) an \emph{LLM-assisted predictor}, HieraMed~\citep{KE2026104087}, which embeds LLM-generated medical-code descriptions; and (4) a \emph{zero-shot LLM}, DeepSeek-V4-Flash~\citep{xu2026deepseek}, evaluated without task-specific parameter updates. Please refer to Appendix~\ref{appendix:baselines} for more details. 

\paragraph{Implementation Details.}
We adopt Qwen3-1.7B~\citep{yang2025qwen3} as the backbone and perform full-parameter fine-tuning throughout all stages. DeepSeek-V4-Flash serves as the teacher model for corpus enrichment. The effective batch sizes for multi-task SID--language alignment, reasoning activation, and reinforcement learning are 1024, 1024, and 256, with learning rates of $3\times10^{-4}$, $1\times10^{-5}$, and $1\times10^{-6}$, respectively. For reinforcement learning, we sample 16 rollouts per input and set the multi-positive loss weight to $\lambda_{\mathrm{MP}}=5\times10^{-3}$. All experiments are conducted on 4 NVIDIA A800 GPUs. Further details are provided in Appendix~\ref{appendix:implementation-details}.
\vspace{-4pt}
\subsection{Main Results (RQ1)}
\label{sec:main-results}
\vspace{-4pt}
\begin{table*}[t]
    \centering
    \caption{Overall next-visit diagnosis prediction performance on MIMIC-III
    and MIMIC-IV. All values are percentages. Best and second-best results are
    shown in bold and underlined, respectively.}
    \label{tab:main-results}
    \normalsize
    \renewcommand{\arraystretch}{1.15}
    \setlength{\tabcolsep}{2.25pt}
    \begin{tabular}{lccccc|ccccc}
        \hline
        Model
        & \multicolumn{5}{c}{MIMIC-III}
        & \multicolumn{5}{c}{MIMIC-IV} \\
        & w-$F_1$ & R@10 & R@20 & R@30 & R@40
        & w-$F_1$ & R@10 & R@20 & R@30 & R@40 \\
        \hline
        RETAIN      & 21.25 & 24.13 & 34.30 & 40.12 & 45.91 &19.54 &25.42 &33.82 &38.91 &42.78 \\
        KAME        & 18.58 & 22.55 & 31.43 & 37.67 & 42.54 &10.59 &17.97 &25.01 &29.56 &33.15 \\
        StageNet    & 18.13 & 21.36 & 30.37 & 36.64 & 41.10 &11.46 &18.41 &25.29 &30.20 &33.82 \\
        CGL       & 21.79 & 25.51 & 35.28 & 41.52 & 46.75 & 20.08&24.07 &33.96 &40.04 &44.31 \\
        TRANS     & 19.10 & 23.32 & 32.26 & 38.39 & 43.14 &15.84 &20.55 &28.71 &34.37 &38.67 \\
        BoxLM     & 22.48 & 25.33 & 35.87 & 41.82 & 46.79 &21.07 &24.15 &34.79 &41.67 &44.53 \\
        HieraMed    & 21.81 & 24.84 & 34.65 & 41.20 & 46.32 &19.10 &24.18 &33.78 &40.87 &44.73 \\
        DeepSeek non-thinking & \underline{30.60} & 25.92 & 36.04 & 40.89 & 43.60 & \textbf{30.22} & 27.81 & 35.66 & 38.90 & 40.20 \\
        DeepSeek thinking & \textbf{30.84} & \underline{27.26} & \underline{37.04} & 42.57 & 46.34 & \underline{27.16} & \underline{29.67}  & 36.88 & 39.71 & 42.45 \\
        \hline
        \method{} non-thinking
        & 25.09 & 26.29 & \underline{37.04}
        & \underline{43.85} & \textbf{50.67} & 24.58 & 28.24 & \underline{37.85} & \underline{43.57} & \underline{48.82} \\
        \method{} thinking
        & 27.58 & \textbf{29.15} & \textbf{39.69}
        & \textbf{46.04} & \underline{50.58} & 26.62 & \textbf{31.49} & \textbf{41.23} & \textbf{46.52} & \textbf{50.19} \\
        \hline
    \end{tabular}
    \vspace{-10pt}
\end{table*}

\paragraph{Performance across both datasets.}
Table~\ref{tab:main-results} shows that \method{} has the highest
w-$F_1$ among EHR-trained models on both datasets and the highest
recall at every reported cutoff across all methods. The thinking
setting leads at seven of eight recall cutoffs; direct inference is
0.09 points higher on MIMIC-III R@40. DeepSeek-V4-Flash has the highest
w-$F_1$ overall on both datasets, and its thinking mode exceeds
\method{}-Think by 0.54 points on MIMIC-IV. Compared with the direct
setting, the thinking model scores 2.49 and 2.04 points higher in
w-$F_1$ on MIMIC-III and MIMIC-IV.

\paragraph{Generalization on MIMIC-IV.}
The MIMIC-IV cohort contains 10,000 sampled patients, and many diagnoses in
its test set do not appear in the EHR training split. All seven EHR-trained
baselines lose 1.41--7.99 points in w-$F_1$ relative to MIMIC-III;
nearly all their recall scores also decline. \method{}-Think loses
0.96 w-$F_1$ points (27.58 to 26.62), but R@10, R@20, and R@30
increase by 2.34, 1.54, and 0.48 points, respectively, while R@40
decreases by 0.39 points. It remains 5.55 w-$F_1$ points above the
strongest EHR-trained baseline, BoxLM, on MIMIC-IV. The smaller w-$F_1$
decline and mixed recall changes are consistent with semantic enrichment
and pretraining supporting prediction under sparse label coverage.
\vspace{-3pt}
\subsection{ICD Ontology Paths in Medical SIDs (RQ2)}
\label{sec:ontology-sid-analysis}
\vspace{-3pt}
\paragraph{Effectiveness of ontology enrichment.}
\vspace{-10pt}
\begin{table}[H]
    \centering
    \caption{SID quality and MIMIC-III recall with disease names alone or
    names plus ICD ontology paths. Entropy is in bits;
    its maximum is 8 for a 256-entry codebook.}
    \label{tab:ontology-sid-comparison}
    \small
    \renewcommand{\arraystretch}{1.08}
    \setlength{\tabcolsep}{7pt}
    \begin{tabular}{llcc}
        \hline
        Metric & Layer & Name only & Name + ICD path \\
        \hline
        R@30 (\%) & & 38.97 & \textbf{39.73} \\
        \hline
        Codebook utilization (\%) & 1 & 100.00 & 100.00 \\
        & 2 & 98.83 & \textbf{100.00} \\
        & 3 & 97.27 & \textbf{100.00} \\
        \hline
        Token distribution entropy & 1 & 7.943 & \textbf{7.951} \\
        & 2 & 7.542 & \textbf{7.919} \\
        & 3 & 7.016 & \textbf{7.736} \\
        \hline
    \end{tabular}
    \vspace{-10pt}
\end{table}

In Table~\ref{tab:ontology-sid-comparison}, the name plus path variant
scores 39.73 R@30 versus 38.97 for name alone after multi-task alignment.
It uses all three codebooks fully. The SID entropy is also
higher in all layers, indicating a more even distribution
across codebook tokens. 
Appendix~\ref{app:sid-semantics} examines the semantic structure of these SIDs through randomly sampled prefix cases and a full-vocabulary comparison of the two variants.
\vspace{-3pt}
\subsection{Ablation Study (RQ3)}
\label{sec:model-study}
\vspace{-3pt}
\paragraph{Effect of Reasoning-Enriched Training for \method{}-Non-think.}
\label{sec:reasoning-enriched-training-nonthink}

\begin{wrapfigure}{r}{0.48\textwidth}
    \centering
    \vspace{-4pt}
    \includegraphics[width=\linewidth]{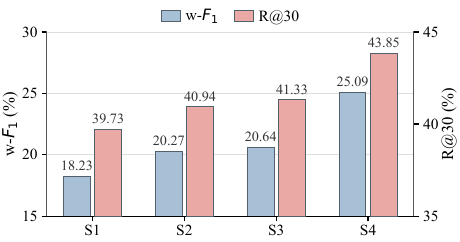}
    \vspace{-2em}
    \caption{Effect of the four stages of reasoning-enriched training on
    MIMIC-III under non-thinking direct inference.}
    \label{fig:reasoning-enriched-training-nonthink}
    \vspace{-10pt}
\end{wrapfigure}

S1 denotes the checkpoint after multi-task SID--clinical context alignment,
before semantic enrichment. Semantic enrichment (S2) raises w-$F_1$ by 2.04
points and R@30 by 1.21 points, showing that SID-grounded narratives add
clinical semantics beyond fixed templates. General corpora (S3) yield small
EHR gains (0.37 w-$F_1$ and 0.39 R@30), but IFEval and GSM8K in
Appendix~\ref{appendix:general-ability} show that they partially recover the
instruction-following and reasoning abilities attenuated during alignment.
Reasoning activation (S4) gives the largest gain over S3 (+4.45 w-$F_1$;
+2.52 R@30), showing that activation improves direct ranking without
test-time reasoning.
\vspace{-3pt}
\paragraph{Effect of the coverage reward for \method{}-Think.}
\label{sec:coverage-aware-training-think}
\vspace{-3pt}
\begin{wrapfigure}{r}{0.48\textwidth}
    \centering
    \vspace{-4pt}
    \includegraphics[width=\linewidth]{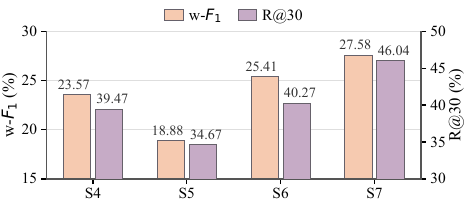}
    \vspace{-2em}
    \caption{Training ablation on MIMIC-III in thinking mode.
    }
    \label{fig:coverage-aware-training-think}
    \vspace{-2em}
\end{wrapfigure}
We compare four settings under the same thinking-mode inference protocol to
separate the effects of visit-level RL and answer-level supervision. 
As shown in Figure~\ref{fig:coverage-aware-training-think}, the
reasoning-activation checkpoint (S4) attains
23.57 w-$F_1$ and 39.47 R@30. With EM reward (S5), every correct rollout
receives equal credit, regardless of how often its SID appears in the group. Its
scores fall to 18.88 and 34.67, respectively. This decline is consistent
with repeated predictions of a few correct diagnoses, a pattern examined
in RQ4. S6 retains S5's RL setup and replaces only the EM reward with the
coverage reward, without adding the multi-positive loss. By discounting
repeated predictions of the same correct SID, S6 reaches 25.41 w-$F_1$ and
40.27 R@30, gains of 6.53 and 5.60 points over S5, respectively, and also
exceeds S4. Adding
the multi-positive loss (S7) raises them to 27.58 and 46.04, gains of
2.17 and 5.77 points over S6. The larger gain in recall is consistent with
the auxiliary loss supplying a learning signal for target diagnoses that the
sampled trajectories may miss. These results support the
complementary roles of rewarding distinct sampled positives and supervising
the full set of target diagnoses.



\vspace{-3pt}
\subsection{Coverage and Diversity in Training (RQ4)}
\label{sec:coverage-diversity}
\vspace{-3pt}
For each visit, we monitor 16 sampled trajectories. Unique SIDs count all
distinct predictions, while unique correct SIDs count distinct next-visit
diagnoses. We also measure mean inverse frequency $1/c_i$ over correct
predictions, where $c_i$ counts repetitions of the predicted SID within the
group; larger values indicate less repetition.

\paragraph{Effect of the coverage reward.}
\begin{figure}[H]
    \centering
    \includegraphics[width=0.85\textwidth]{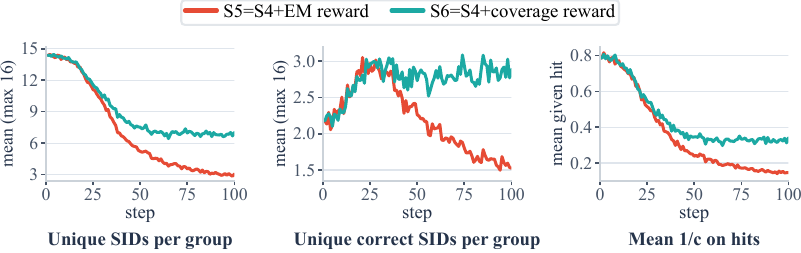}
    \vspace{-1em}
    \caption{MIMIC-III trajectories for S5 (EM reward) and S6 (coverage reward), both initialized from S4.}
    \label{fig:coverage-reward-diversity}
    \vspace{-1em}
\end{figure}

Under EM reward, unique correct SIDs rise from 2.18 at step 1 to 2.86 at
step 30, then fall to 1.52 by step 100 (Figure~\ref{fig:coverage-reward-diversity}).
The coverage reward retains 2.89 unique correct SIDs at step 100 and produces
7.03 unique SIDs per group, versus 3.06 under EM. S6's higher mean inverse
frequency indicates less repetition among correct predictions, consistent
with S6's higher held-out scores in Figure~\ref{fig:coverage-aware-training-think}.

\paragraph{Effect of the multi-positive loss.}
\begin{figure}[H]
    \centering
    \includegraphics[width=0.85\textwidth]{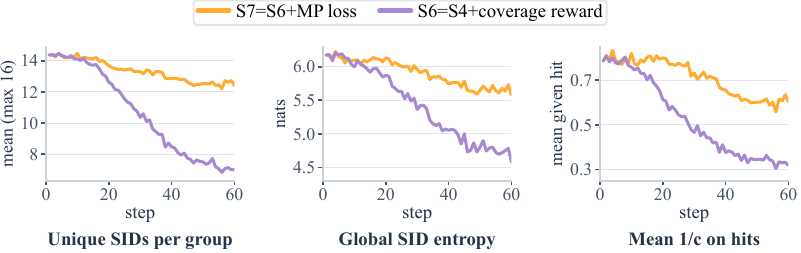}
    \vspace{-1em}
    \caption{MIMIC-III trajectories for S6 and S7. The missing S7 value at step 42 is linearly interpolated between steps 41 and 43.}
    \label{fig:mp-loss-diversity}
    \vspace{-1em}
\end{figure}

The coverage reward can distinguish correct SIDs only after sampling.
At step 60, S7 produces 12.40 unique SIDs per group, versus 7.02 for S6
(Figure~\ref{fig:mp-loss-diversity}). Batch-level SID entropy,
$-\sum_s p(s)\log p(s)$ for empirical frequencies $p(s)$ of parsed final SIDs,
is 5.58 versus 4.58, and mean inverse frequency
among correct predictions is 0.60 versus 0.32. The additional supervision
thus reduces repetition while retaining more candidate diagnoses; S7 also
improves held-out w-$F_1$ and R@30 over S6
(Figure~\ref{fig:coverage-aware-training-think}).




\vspace{-2pt}
\section{Conclusion}
\vspace{-2pt}
Predicting diagnoses at the next visit requires a model to recover several
valid diagnoses from a large vocabulary. Rewarding each reasoning trajectory
only for correct predictions can favor repeated answers and leave other
diagnoses uncovered. We introduced \method{}, which represents diagnoses
with compositional medical SIDs grounded in clinical language and patient
histories. The model learns to reason before predicting a SID, while a
coverage reward and supervision over multiple correct diagnoses encourage
complementary answers across trajectories. It supports direct ranking and
multiple reasoning chains. On MIMIC-III and MIMIC-IV, \method{} reaches
R@30 of 46.04\% and 46.52\%, respectively, has the highest recall at
every reported cutoff, and exceeds all EHR-trained baselines in weighted $F_1$.
We discuss broader impacts, limitations, and future work in
Appendix~\ref{appendix:broader-impacts-limitations-future-work}.

\providecommand{\eg}{\emph{e.g., }}

\subsection*{AI use statement}

We primarily utilized LLMs for paper polishing and programming assistance.
Specifically, for paper writing, we employed LLMs (\eg GPT-5.6) to refine sentence fluency as well as provide suggestions on precise and appropriate expression.
All the modifications generated by LLMs were carefully reviewed by the authors to ensure that the final content faithfully reflects the authors' intended ideas.
For code generation, LLMs (\eg GPT-5.6) were mainly used to generate some code snippets, which were subsequently adapted and integrated into the implementation framework by the authors.
In both paper polishing and programming assistance, the authors were in full control, with LLMs serving solely as tools to improve efficiency.

\subsection*{Reproducibility statement}

For reproducibility, we describe the method and training objectives in
Section~\ref{sec:method} and Appendix~\ref{appendix:task-formats}.
The coverage reward property is proved in Appendix~\ref{appendix:coverage-reward-property}. Data preprocessing,
implementation details, and experimental settings are provided in
Appendix~\ref{appendix:experimental-details}. The code, training logs,
and test logs are available at
\url{https://github.com/zmlxzyh/CARing-Codes-Logs}.

\bibliography{iclr2027_conference}
\bibliographystyle{iclr2027_conference}

\clearpage
\appendix
\let\addcontentsline\appendixaddcontentsline
\renewcommand{\contentsname}{Appendix Contents}
\setcounter{tocdepth}{3}
\tableofcontents
\clearpage

\section{Extended Related Work}
\label{appendix:related-work}

We expand the discussion in Section~\ref{sec:related-work}, following the
progression from clinical prediction and medical representations to reasoning
models and the optimization of diagnosis coverage.

\subsection{Longitudinal Diagnosis Prediction}

Longitudinal EHR models learn how diagnoses and other clinical events evolve
across visits. Doctor AI and Dipole use recurrent networks to model these
sequences, while RETAIN uses attention to identify influential visits and
clinical variables~\citep{choi2016doctor,ma2017dipole,choi2016retain}.
HiTANet accounts for irregular time intervals, and StageNet models changes in
patient status~\citep{luo2020hitanet,gao2020stagenet}.

Medical knowledge and relations between patients provide additional structure.
GRAM and KAME incorporate medical ontologies into the representations used for
prediction~\citep{choi2017gram,ma2018kame}. CGL learns from patient and disease
graphs, while subsequent methods model transitions on dynamic disease graphs
or encode EHRs as temporal heterogeneous graphs
\citep{lu2021collaborative,lu2022context,chen2024predictive}.
NECHO combines medical codes, demographic information, and clinical notes
through contrastive learning and hierarchical regularization
\citep{koo2024next}. BoxLM combines the textual semantics and hierarchical
structure of medical concepts using box embeddings~\citep{tan2025boxlm}.
HieraMed uses embeddings of medical code descriptions within a hierarchy of
code, visit, and patient representations~\citep{KE2026104087}.

Pretraining on patient records offers another route to clinical prediction.
BEHRT and Med-BERT learn contextual representations of structured EHRs that
can be adapted to disease prediction~\citep{li2020behrt,rasmy2021med}.
Generative approaches also exist: TransformEHR predicts future medical codes
with a Transformer encoder and decoder, and Foresight models patient timelines
through autoregressive generation
\citep{yang2023transformehr,kraljevic2024foresight}.
CPLLM adapts pretrained LLMs to clinical prediction by expressing patient
histories as text, with evaluations including specific disease prediction and
readmission~\citep{benshoham2024cpllm}. These settings differ in the information
available to the model and the outputs being evaluated. Predicting a selected
outcome, generating future clinical concepts, and recovering the full set of
diagnoses at the next visit place different demands on the prediction model.

\subsection{Medical Code Representations and Generation}

Medical codes identify standardized concepts, but their character sequences do
not express disease meanings in ordinary language. Evaluations of LLMs on
medical code queries report errors in mapping descriptions to exact codes
\citep{soroush2024large}. This motivates representations that connect code
identities to their descriptions and clinical relations.
MedTok encodes textual descriptions and relational context, then quantizes
these representations into a shared token space~\citep{su2025multimodal}.
Its evaluations include both EHR predictors and medical question answering,
where the learned tokens are supplied as prefixes to LLMs.
MedRQ uses residual quantization and supervision from medical ontologies to
learn discrete medical representations~\citep{lv2026towards}.

MERA incorporates disease identities directly into a language model's
vocabulary, with a dedicated token for each disease~\citep{ma2025memorize}.
It learns mappings between disease tokens and definitions, varies the order
of diagnoses within a visit, and uses the medical hierarchy in candidate
ranking. This treatment accounts for the unordered structure of visit
diagnoses, although a dedicated token for each disease requires the added
vocabulary to grow with the diagnosis catalog.

Our compositional representation draws on semantic IDs in generative
recommendation, including systems that generate a natural language rationale
before an identifier
\citep{rajput2023recommender,liu2025onerec, he2026reasoning, team2026onereason}.
We adapt this representation to diseases by quantizing their descriptions and
ICD ontology paths into sequences of shared codebook tokens. Alignment with
clinical text and patient histories makes these SIDs usable as both inputs
and diagnosis outputs. The resulting interface allows a single autoregressive
model to generate a clinical rationale followed by a diagnosis SID.

\subsection{Reasoning for Clinical Prediction}

Medical question answering has provided a setting for adapting and evaluating
LLM reasoning. Med-PaLM and Med-PaLM~2 study the use of LLMs for answering
medical questions~\citep{singhal2023large,singhal2025toward}.
HuatuoGPT-o1 trains on medical problems with verifiable answers, while
AlphaMed studies reinforcement learning from answer correctness without an
initial stage of supervised training on reasoning traces
\citep{chen2024huatuogpt,liu2025beyond}.
The m1 study examines computation at inference time for medical reasoning and
finds that longer reasoning alone does not consistently improve performance;
medical knowledge and the reasoning budget also matter~\citep{huang2025m1}.

For prediction from patient records, knowledge retrieval and reasoning
supervision help connect clinical evidence to target outcomes.
GraphCare constructs patient-specific knowledge graphs for healthcare
prediction~\citep{jiang2024graphcare}.
KARE retrieves relevant knowledge graph communities and distills reasoning
for mortality and readmission prediction~\citep{jiang2025reasoning}.
ReMedi uses known outcomes to guide the construction and regeneration of
rationale--answer pairs for supervised and preference tuning
\citep{cao2026remedi}. EAG-RL initializes the policy with reasoning trajectories
guided by specialized EHR models. It then uses their attention patterns as an
additional reward during reinforcement learning~\citep{fang2026toward}.

Reasoning has also been applied to medical coding. DeepICD-R1 combines
distilled reasoning traces with rewards at several levels of the ICD hierarchy
to predict codes from admission notes~\citep{rohr2026deepicd}.
Coding the conditions described in a clinical document and forecasting a
future visit use different evidence and supervision. Our setting requires
reasoning from earlier visits to recover multiple diagnoses in the next
visit. We therefore evaluate diagnosis coverage across sampled trajectories,
in addition to whether an individual trajectory predicts a correct diagnosis.

\subsection{Large Reasoning Models and Policy Collapse}

Chain-of-thought prompting elicits intermediate reasoning steps before an
answer~\citep{wei2022chain}. Self-consistency samples several reasoning paths
and aggregates their answers~\citep{wang2023self}, while work on scaling
computation at inference time studies how to allocate search and refinement
to different problems~\citep{snell2024scaling}.
Reinforcement learning with verifiable rewards trains reasoning policies from
feedback on their outputs. DeepSeekMath introduces GRPO, which compares
sampled trajectories within a group, and DeepSeek-R1 applies reinforcement
learning to develop reasoning capabilities
\citep{shao2024deepseekmath,guo2025deepseek}.

Higher accuracy for an individual sample need not imply broader coverage
across samples. In the settings evaluated by \citet{yue2025does}, RLVR improves
pass@1, but base models can attain higher pass@$k$ when $k$ is large.
\citet{gxchen2025kl} analyze how rewards, the reference policy, and KL
regularization determine the target distribution. Under conditions including
weak regularization and equal verifiable rewards, that distribution can
concentrate on a single region of high reward. These findings motivate
examining both correctness and coverage when several outputs can be valid.
For diagnosis prediction, repeated success on one diagnosis can coexist with
poor recall of the other diagnoses recorded at the same visit.

\subsection{Correctness and Diversity in Reinforcement Learning}

Optimizing the quality of a group of outputs requires assigning credit for
what each sample contributes to that group. GAPO adjusts rewards using the
frequency of valid outputs within a rollout group, giving more credit to
underrepresented valid outputs~\citep{anschel2025group}.
SetPO measures diversity through similarities between trajectories and uses
each trajectory's marginal contribution to group diversity in policy
optimization~\citep{li2026setpo}.
These methods connect the training signal for one trajectory to the other
trajectories sampled for the same input.

Rewards defined over sets also appear in multi-label classification.
\citet{yang2019deep} use reinforcement learning with a reward independent of
label order to reduce the sensitivity of sequence generation to arbitrary
label orderings. Their model generates a label set within one output
sequence. In our formulation, each reasoning trajectory ends with one
diagnosis, and the rollout group is evaluated against the complete set of
diagnoses at the next visit.

Our objective measures diversity in terms of distinct correct diagnosis
labels. The coverage reward divides a correct prediction's credit by the
number of times that diagnosis appears in the group. Consequently, each
distinct correct diagnosis contributes one unit to the raw group reward,
as shown in Appendix~\ref{appendix:coverage-reward-property}.
This gives the frequency adjustment a direct interpretation as diagnosis
coverage. An auxiliary multi-positive loss supervises the SID answer tokens
to provide training signal for correct diagnoses absent from a sampled
group. Together, the two terms address repetition among sampled positives
and the limited exposure to positives that sampling can miss.

\section{Additional Method Details}
\label{appendix:task-formats}

This section provides representative input--output examples for the task
formats used throughout training. We first present the six tasks used for
multi-task SID--language alignment. The examples below follow the prompts in
the implementation; only the assistant response is supervised.

\makeatletter
\@ifundefined{sidtask}{%
  \newtcolorbox{sidtask}[1]{%
    colback=gray!5,
    colframe=black!60,
    boxrule=0.5pt,
    arc=2pt,
    left=6pt,
    right=6pt,
    top=6pt,
    bottom=6pt,
    title={#1},
    fonttitle=\bfseries\small,
    coltitle=white,
    colbacktitle=black!70,
    before skip=8pt,
    after skip=8pt,
    before upper={\raggedright\sloppy}%
  }%
}{}
\@ifundefined{sidprompt}{%
  \newtcolorbox{sidprompt}[1]{%
    enhanced,
    breakable,
    colback=gray!5,
    colframe=black!60,
    boxrule=0.5pt,
    arc=2pt,
    left=6pt,
    right=6pt,
    top=6pt,
    bottom=6pt,
    title={#1},
    title after break={#1 (continued)},
    fonttitle=\bfseries\small,
    coltitle=white,
    colbacktitle=black!70,
    before skip=8pt,
    after skip=8pt,
    before upper={\raggedright\sloppy}%
  }%
}{}
\makeatother

\newcommand{\sidcode}[1]{\texttt{\detokenize{#1}}}

\subsection{Multi-Task SID--Language Alignment}
\label{appendix:multitask-task-formats}

Tasks T1--T4 align textual or SID-based longitudinal
histories with disease-name or SID prediction. Tasks T5--T6 establish bidirectional grounding between disease names and compositional SIDs. Diagnoses within each visit are
unordered, whereas visits are listed chronologically.

Let $\mathcal{D}_{\tau}$ denote the training set for task $\tau\in\{1,\ldots,6\}$ and let $\mathcal{A}(y)$ contain the assistant-response token positions. 
All tasks are jointly optimized under the standard next-token prediction objective.
\begin{equation}
    \mathcal{L}_{\mathrm{MT}}
    =-\frac{1}{\sum_{\tau}|\mathcal{D}_{\tau}|}
    \sum_{\tau=1}^{6}\sum_{(x,y)\in\mathcal{D}_{\tau}}
    \sum_{j\in\mathcal{A}(y)}
    \log p_{\theta}(y_j\mid x,y_{<j}).
    \label{eq:multitask-alignment}
\end{equation}

\begin{sidtask}{Task T1: SID History to SID}
  \small
  \textbf{System:}\\
  \textit{Predict one possible diagnosis SID in the next visit from the
  chronological SID history.}\\
  \textbf{User:}\\
  The patient's chronological diagnosis history is:\\
  \sidcode{<visit_start>}\\
  \sidcode{<a_117><b_59><c_117>},
  \sidcode{<a_199><b_76><c_251>}\\
  \sidcode{<visit_end>}\\
  \sidcode{<visit_start>}\\
  \sidcode{<a_117><b_0><c_82>},
  \sidcode{<a_87><b_184><c_39>}\\
  \sidcode{<visit_end>}\\
  Return exactly one disease SID.\\
  \textbf{Assistant:}\\
  \sidcode{<a_117><b_0><c_82>}
\end{sidtask}

\begin{sidtask}{Task T2: Textual History to SID}
  \small
  \textbf{System:}\\
  \textit{Predict one possible diagnosis SID in the next visit from the
  chronological diagnosis history.}\\
  \textbf{User:}\\
  The patient's chronological diagnosis history is:\\
  Visit 1: Subarachnoid hemorrhage; Pneumonitis due to inhalation of food or
  vomitus\\
  Visit 2: Cerebral aneurysm, nonruptured; Unspecified essential hypertension\\
  Return exactly one disease SID.\\
  \textbf{Assistant:}\\
  \sidcode{<a_117><b_0><c_82>}
\end{sidtask}

\begin{sidtask}{Task T3: SID History to Disease Name}
  \small
  \textbf{System:}\\
  \textit{Predict one possible diagnosis in the next visit from the
  chronological SID history.}\\
  \textbf{User:}\\
  The patient's chronological diagnosis history is:\\
  \sidcode{<visit_start>}\\
  \sidcode{<a_117><b_59><c_117>},
  \sidcode{<a_199><b_76><c_251>}\\
  \sidcode{<visit_end>}\\
  \sidcode{<visit_start>}\\
  \sidcode{<a_117><b_0><c_82>},
  \sidcode{<a_87><b_184><c_39>}\\
  \sidcode{<visit_end>}\\
  Return one possible disease description for the next visit.\\
  \textbf{Assistant:}\\
  \textit{Cerebral aneurysm, nonruptured}
\end{sidtask}

\begin{sidtask}{Task T4: Textual History to Disease Name}
  \small
  \textbf{System:}\\
  \textit{Predict one possible diagnosis in the next visit from the
  chronological diagnosis history.}\\
  \textbf{User:}\\
  The patient's chronological diagnosis history is:\\
  Visit 1: Subarachnoid hemorrhage; Pneumonitis due to inhalation of food or
  vomitus\\
  Visit 2: Cerebral aneurysm, nonruptured; Unspecified essential hypertension\\
  Return one possible disease description for the next visit.\\
  \textbf{Assistant:}\\
  \textit{Cerebral aneurysm, nonruptured}
\end{sidtask}

\begin{sidtask}{Task T5: Disease Name to SID}
  \small
  \textbf{System:}\\
  \textit{You map ICD disease names to semantic identifiers.}\\
  \textbf{User:}\\
  Which disease has the name ``Cerebral aneurysm, nonruptured''?\\
  \textbf{Assistant:}\\
  \sidcode{<a_117><b_0><c_82>}
\end{sidtask}

\begin{sidtask}{Task T6: SID to Disease Name}
  \small
  \textbf{System:}\\
  \textit{You map semantic identifiers to ICD disease names.}\\
  \textbf{User:}\\
  What disease does \sidcode{<a_117><b_0><c_82>} represent?\\
  \textbf{Assistant:}\\
  \textit{Cerebral aneurysm, nonruptured}
\end{sidtask}

\subsection{Reasoning-Enriched Training}
\label{appendix:reasoning-enriched-task-formats}

The disease-level and case-level enriched corpora are trained as plain-text
next-token prediction sequences, without a system or user prompt. Reasoning
activation instead uses the chat-formatted history-only prompt shown in the
third example; both its reasoning trace and final SID are supervised.

\begin{sidtask}{Training Case R1: Disease-Level Semantic Enrichment}
  \small
  \textbf{Training text:}\\
  \sidcode{<a_117><b_0><c_82>} is a clinically significant vascular
  abnormality characterized by localized ballooning of an intracranial artery
  that has not yet ruptured. The CCI body-system classification places this
  entity within Diseases of the circulatory system, and the CCI chronicity
  classification designates it as chronic, reflecting its persistent nature
  and need for long-term surveillance. According to the CCS single-level
  category, \sidcode{<a_117><b_0><c_82>} falls under Other and ill-defined
  cerebrovascular disease, and the complete CCS multi-level hierarchy is
  Diseases of the circulatory system $>$ Cerebrovascular disease $>$ Other and
  ill-defined cerebrovascular disease [111.]. Within the ICD ontology
  hierarchy, it is situated as DISEASES AND INJURIES $>$ DISEASES OF THE
  CIRCULATORY SYSTEM $>$ CEREBROVASCULAR DISEASE $>$ Other and ill-defined
  cerebrovascular disease $>$ \sidcode{<a_117><b_0><c_82>}. This classification
  structure is coding information and should not be interpreted as evidence of
  a causal pathway between the categories and the lesion. Clinically,
  \sidcode{<a_117><b_0><c_82>} often remains asymptomatic and may be discovered
  incidentally during brain imaging for unrelated complaints. However, the
  most feared complication is rupture, leading to subarachnoid hemorrhage,
  which carries high morbidity and mortality. Established risk factors for the
  development and growth of \sidcode{<a_117><b_0><c_82>} include hypertension,
  cigarette smoking, alcohol misuse, and a family history of intracranial
  aneurysms. Certain genetic conditions, such as autosomal dominant polycystic
  kidney disease and Ehlers-Danlos syndrome, also predispose patients to this
  vascular lesion. Common comorbidities among patients harboring
  \sidcode{<a_117><b_0><c_82>} may include atherosclerotic cardiovascular
  disease, prior ischemic stroke, or other manifestations of systemic vascular
  injury. Management strategies focus on preventing rupture, balancing the
  risks of microsurgical clipping or endovascular coiling against the natural
  history of the unruptured lesion. Regular imaging follow-up is often
  recommended for small stable lesions, while larger or symptomatic aneurysms
  may warrant intervention. Thus, \sidcode{<a_117><b_0><c_82>} represents a
  chronic circulatory condition that demands individualized risk assessment
  and careful clinical surveillance.
\end{sidtask}

\begin{sidtask}{Training Case R2: Case-Level Semantic Enrichment}
  \small
  \textbf{Training text:}\\
  Reviewing the longitudinal record, the picture centers on chronic
  cerebrovascular disease. At the initial visit,
  \sidcode{<a_117><b_0><c_82>} appeared alongside
  \sidcode{<a_117><b_59><c_117>}, with associated symptoms
  \sidcode{<a_138><b_32><c_38>} and \sidcode{<a_93><b_60><c_3>};
  \sidcode{<a_179><b_208><c_18>} was also present but seemed less tied to this
  axis. By the most recent visit, recurrence of
  \sidcode{<a_117><b_59><c_117>} together with
  \sidcode{<a_207><b_3><c_118>} and \sidcode{<a_38><b_213><c_64>} points to an
  active cerebrovascular process that required intervention and now
  device-related follow-up. In this context, the chronic
  \sidcode{<a_117><b_0><c_82>} from the earliest encounter stands out as the
  likely underlying lesion being monitored, rather than an incidental finding,
  since it belongs to the same circulatory/cerebrovascular domain. The early
  presence of \sidcode{<a_179><b_208><c_18>} contributed to the initial burden
  but does not share this close relationship. Going forward, continued
  cerebrovascular surveillance and management of potential complications,
  along with related neurological and vascular diagnoses, would be expected.
\end{sidtask}

\begin{sidtask}{Training Case R3: Reasoning Activation}
  \small
  \textbf{System:}\\
  \textit{Predict one possible diagnosis SID in the next visit from the
  chronological SID history. Think through the longitudinal clinical patterns
  before answering.}\\
  \textbf{User:}\\
  The patient's chronological diagnosis history is:\\
  \sidcode{<visit_start>}\\
  \sidcode{<a_117><b_0><c_82>},
  \sidcode{<a_138><b_32><c_38>},
  \sidcode{<a_117><b_59><c_117>},
  \sidcode{<a_179><b_208><c_18>},
  \sidcode{<a_93><b_60><c_3>}\\
  \sidcode{<visit_end>}\\
  \sidcode{<visit_start>}\\
  \sidcode{<a_117><b_59><c_117>},
  \sidcode{<a_207><b_3><c_118>},
  \sidcode{<a_38><b_213><c_64>}\\
  \sidcode{<visit_end>}\\
  Think step by step about the longitudinal clinical patterns, then return
  exactly one disease SID.\\
  \textbf{Assistant:}\\
  \sidcode{<think>}\\
  I see a core longitudinal pattern centered on chronic cerebrovascular
  disease. The earliest visit already contained
  \sidcode{<a_117><b_0><c_82>}, a chronic circulatory-system diagnosis,
  alongside \sidcode{<a_117><b_59><c_117>} and symptoms such as
  \sidcode{<a_138><b_32><c_38>} and \sidcode{<a_93><b_60><c_3>}. The
  recurrence of \sidcode{<a_117><b_59><c_117>} in the most recent visit,
  together with \sidcode{<a_207><b_3><c_118>} and
  \sidcode{<a_38><b_213><c_64>}, strongly suggests an active cerebrovascular
  process that required intervention and generated device-related follow-up.
  The persistence of the chronic SID \sidcode{<a_117><b_0><c_82>} from the
  first visit may remain clinically relevant because it belongs to the same
  circulatory/cerebrovascular domain and could represent the underlying lesion
  being monitored. The non-chronic \sidcode{<a_179><b_208><c_18>} from the
  first visit seems less tied to this axis, though it contributed to the
  initial overall burden. Given the recurrence and the recent device/procedure
  encounters, subsequent visits may continue to involve cerebrovascular
  surveillance, complication management, or related neurological and vascular
  diagnoses.\\
  \sidcode{</think>}\\[0.4em]
  \sidcode{<a_117><b_0><c_82>}
\end{sidtask}

\subsubsection{Training Objectives}
\label{appendix:reasoning-enriched-objectives}

Let $\mathcal{X}_{\mathrm{disease}}$ and $\mathcal{X}_{\mathrm{case}}$ denote the
disease-level and case-level enriched corpora. Each sample
$x=(x_1,\ldots,x_{|x|})$ is a plain-text sequence, with
$p_\theta(x)=\prod_{j=1}^{|x|}p_\theta(x_j\mid x_{<j})$. The two corpora use
next-token prediction objectives:
\begin{equation}
\begin{aligned}
    \mathcal{L}_{\mathrm{disease}}
    &=-\mathbb{E}_{x\sim\mathcal{X}_{\mathrm{disease}}}
      \left[\log p_\theta(x)\right],\\
    \mathcal{L}_{\mathrm{case}}
    &=-\mathbb{E}_{x\sim\mathcal{X}_{\mathrm{case}}}
      \left[\log p_\theta(x)\right].
\end{aligned}
\label{eq:reasoning-enrichment-losses}
\end{equation}

For reasoning activation, each training example pairs a historical SID
sequence $H$ with a teacher-generated reasoning path $r$ and a target
diagnosis $d$. The prompt contains only $H$, and the loss applies to the
assistant response:
\begin{equation}
    \mathcal{L}_{\mathrm{act}}
    =-\mathbb{E}_{(H,r,d)}\!\left[
    \log p_{\theta}\!\left(\texttt{<think>}r\texttt{</think>}\,q(d)\mid H\right)
    \right].
    \label{eq:reasoning-activation}
\end{equation}

\subsection{Prompts for Reasoning-Enriched Data}
\label{appendix:reasoning-enriched-data-prompts}

We generate the three supervision signals in two routes. Disease-level
semantic enrichment directly converts the structured dossier of one coded
concept into a SID-grounded clinical narrative. For case-level enrichment, we
first elicit a history-supported reasoning path and then provide that path,
together with the same history and the observed diagnosis, to a separate
integration prompt. The resulting reasoning path is also reused as the
reasoning supervision in the activation data. The prompts below omit only the
mechanics of model invocation.

\begin{sidprompt}{Prompt P1: Disease-Level Narrative Generation}
  \small
  \textbf{User prompt (template):}\\
  You are a senior clinical medicine expert preparing an in-depth narrative
  for a coded clinical concept dossier.\\[0.3em]
  \textbf{Source information:}\\
  -- Concept name, provided only for understanding and medical knowledge
  retrieval; never reproduce it in the output: [CONCEPT NAME]\\
  -- Semantic Identifier: [SID]\\
  -- CCI body-system classification: [CCI BODY SYSTEM]\\
  -- CCI chronicity classification: [CCI CHRONICITY]\\
  -- CCS single-level category: [CCS SINGLE-LEVEL LABEL]\\
  -- CCS optional category: [CCS OPTIONAL LABEL]\\
  -- CCS multi-level hierarchy: [COMPLETE CCS MULTI-LEVEL PATH]\\
  -- ICD ontology hierarchy: [COMPLETE ICD ONTOLOGY PATH]\\[0.3em]
  \textbf{Task:}\\
  1. Write a rich, coherent clinical narrative of at least ten sentences about
  the coded clinical concept represented by [SID].\\
  2. Incorporate every available structured fact above. Explicitly describe
  the CCI body-system classification, CCI chronicity classification, CCS
  single-level category, complete CCS multi-level hierarchy, and complete ICD
  ontology hierarchy.\\
  3. Treat the CCI, CCS, and ICD fields as coding and classification
  information. Do not present a classification relationship as proof of
  clinical causation.\\
  4. Enrich the dossier using only widely accepted, high-confidence clinical
  knowledge relevant to the supplied concept name. When applicable, discuss a
  concise clinical description, established causes or risk factors, commonly
  associated complications or downstream clinical consequences, relevant
  comorbidities or prior diagnoses, and the broader clinical context in which
  the concept may be recorded.\\
  5. If CCI chronicity or body-system information is unavailable, state only
  that the concept is not classified for that field. Do not infer a missing
  classification.\\
  6. Adapt the narrative to the type of coded concept. If it represents an
  external cause, status, symptom, finding, or supplementary classification
  rather than a disease, do not describe it as a disease.\\[0.3em]
  \textbf{Entity-reference requirements:}\\
  7. Begin the paragraph with [SID].\\
  8. Every direct reference to this specific clinical concept must use exactly
  [SID]. Never output the supplied concept name, ICD code, disease ID,
  abbreviation, synonym, or another alias.\\
  9. Official CCS, CCI, and ICD category labels supplied above may appear only
  when explicitly describing those classification categories.\\
  10. Use [SID] naturally at least four times. Do not output any other Semantic
  Identifier.\\[0.3em]
  \textbf{Output requirements:}\\
  11. Output only one flowing English paragraph. Do not use headings, bullet
  points, enumerations, JSON, citations, meta-commentary, or
  \sidcode{<think>} tags.\\
  12. Maintain a professional and medically precise tone suitable for a
  clinical concept dossier.\\[0.3em]
  The concrete output obtained by instantiating this template with
  \sidcode{<a_117><b_0><c_82>} is shown in Training Case R1.
\end{sidprompt}

\begin{sidprompt}{Prompt P2: History-Supported Reasoning-Path Generation}
  \small
  \textbf{User prompt (template):}\\
  You are an experienced clinical diagnostician reviewing a patient's
  longitudinal diagnosis history.\\[0.3em]
  Your task is to reason through the patient's historical diagnoses and
  identify the longitudinal clinical patterns that would be most useful for a
  downstream prediction of one possible diagnosis in the next visit. The
  observed next diagnosis is provided only as internal guidance for deciding
  which historical patterns may be relevant. Never mention or discuss that
  diagnosis.\\[0.3em]
  \textbf{Patient diagnosis history:}\\
  Visits are listed in chronological order, from oldest to most recent.
  Diagnoses within the same visit are an unordered set; never infer temporal
  order among them. For every diagnosis, the input supplies its complete SID,
  diagnosis name, CCI body system, CCI chronicity, CCS single-level category,
  and CCS multi-level path.\\
  \mbox{[HISTORICAL VISITS WITH PER-DIAGNOSIS METADATA]}\\[0.3em]
  \textbf{Observed next diagnosis---internal guidance only; do not reveal:}\\
  \mbox{[NEXT-DIAGNOSIS SID, NAME, CCI FIELDS, AND CCS FIELDS]}\\[0.3em]
  \textbf{SID atomicity rule:} Every SID is one atomic identifier consisting
  of exactly three adjacent components,
  \sidcode{<a_NUMBER><b_NUMBER><c_NUMBER>}. Copy the complete SID exactly;
  never abbreviate it to one or two components.\\[0.3em]
  \textbf{Output requirements:}\\
  1. Output only a concise but detailed first-person reasoning monologue as one
  paragraph. Start directly with the reasoning.\\
  2. Compare diagnoses across visits, while treating diagnoses within one visit
  as unordered and potentially co-occurring.\\
  3. Give the greatest immediate predictive weight to recent visits. Also
  consider diagnoses that persist or recur across visits, especially chronic
  diagnoses. For a single-visit history, discuss only the observed disease
  burden.\\
  4. Body systems, chronicity, CCS categories, and broad clinical relationships
  may be used as attributes of SID-based diagnoses, but never as alternative
  names for a concrete diagnosis.\\
  5. Whenever referring to a concrete diagnosis, use only its complete SID.
  Never attach a diagnosis name, abbreviation, synonym, clinical gloss, or
  explanatory label to a SID.\\
  6. Mention only SIDs present in the history. Never introduce an unseen SID,
  and do not output the observed next-diagnosis SID unless it already occurs in
  the history.\\
  7. Never mention ``observed diagnosis,'' ``reference diagnosis,'' ``target
  diagnosis,'' or equivalent language. Reason as if predicting blindly.\\
  8. Do not claim that chronology or within-visit co-occurrence proves
  causation. Use appropriately cautious language such as ``may,'' ``could,''
  or ``is consistent with.''\\
  9. Adapt the depth to history length: emphasize burden and co-occurrence for
  short histories, and persistence, recurrence, newly appearing conditions,
  and recent shifts for longer histories.\\
  10. Do not select or name a specific next diagnosis. End with a
  non-deterministic summary.\\[0.3em]
  Applied to the two-visit history in Training Case R3, this prompt produces
  the paragraph enclosed by \sidcode{<think>} and \sidcode{</think>} in that
  example. That paragraph is retained as activation supervision and passed to
  Prompt P3 as the case-level reasoning path.
\end{sidprompt}

\begin{sidprompt}{Prompt P3: Case-Level Narrative Integration}
  \small
  \textbf{User prompt (instantiated example):}\\
  Integrate the following information into a single, coherent, natural
  narrative paragraph.\\[0.3em]
  \textbf{Patient diagnosis history:}\\
  Visits are listed in chronological order, from oldest to most recent.
  Diagnoses within the same visit are an unordered set.\\
  Historical visit 1---oldest:\\
  -- \sidcode{<a_117><b_0><c_82>}: Cerebral aneurysm, nonruptured\\
  -- \sidcode{<a_138><b_32><c_38>}: Headache\\
  -- \sidcode{<a_117><b_59><c_117>}: Subarachnoid hemorrhage\\
  -- \sidcode{<a_179><b_208><c_18>}: Iron deficiency anemia, unspecified\\
  -- \sidcode{<a_93><b_60><c_3>}: Visual discomfort\\
  Historical visit 2---most recent:\\
  -- \sidcode{<a_117><b_59><c_117>}: Subarachnoid hemorrhage\\
  -- \sidcode{<a_207><b_3><c_118>}: Other complications due to nervous system
  device, implant, and graft\\
  -- \sidcode{<a_38><b_213><c_64>}: Other specified procedures as the cause of
  abnormal reaction or later complication\\[0.3em]
  \textbf{Observed next diagnosis to integrate:}\\
  -- SID: \sidcode{<a_117><b_0><c_82>}\\
  -- Diagnosis name: Cerebral aneurysm, nonruptured\\[0.3em]
  \textbf{Reasoning path:}\\
  I see a core longitudinal pattern centered on chronic cerebrovascular
  disease. The earliest visit already contained
  \sidcode{<a_117><b_0><c_82>}, a chronic circulatory-system diagnosis,
  alongside \sidcode{<a_117><b_59><c_117>} and symptoms such as
  \sidcode{<a_138><b_32><c_38>} and \sidcode{<a_93><b_60><c_3>}. The
  recurrence of \sidcode{<a_117><b_59><c_117>} in the most recent visit,
  together with \sidcode{<a_207><b_3><c_118>} and
  \sidcode{<a_38><b_213><c_64>}, strongly suggests an active cerebrovascular
  process that required intervention and generated device-related follow-up.
  The persistence of the chronic SID \sidcode{<a_117><b_0><c_82>} from the
  first visit may remain clinically relevant because it belongs to the same
  circulatory/cerebrovascular domain and could represent the underlying lesion
  being monitored. The non-chronic \sidcode{<a_179><b_208><c_18>} from the
  first visit seems less tied to this axis, though it contributed to the
  initial overall burden. Given the recurrence and the recent device/procedure
  encounters, subsequent visits may continue to involve cerebrovascular
  surveillance, complication management, or related neurological and vascular
  diagnoses.\\[0.3em]
  \textbf{Output requirements:}\\
  1. Write one natural, flowing longitudinal clinical narrative. Avoid
  mechanical or formulaic language.\\
  2. Preserve the essential analytical insights and clinical logic from the
  reasoning path.\\
  3. Connect the historical pattern to the observed SID as an ordinary clinical
  finding. Never use meta-language such as ``reference,'' ``target,'' ``ground
  truth,'' or ``next diagnosis,'' and never imply that it was known in advance.\\
  4. Refer to every concrete diagnosis only by its complete SID of the form
  \sidcode{<a_XXX><b_YYY><c_ZZZ>}. Do not output diagnosis names, codes,
  abbreviations, synonyms, or parenthetical clinical glosses. Use only SIDs
  appearing in the supplied history or observed diagnosis.\\
  5. Keep the narrative clinically precise, cohesive, and engaging. Output only
  the paragraph.\\[0.3em]
  The narrative produced for this instantiated prompt is shown in Training
  Case R2.
\end{sidprompt}

\subsection{Coverage Reward Property}
\label{appendix:coverage-reward-property}

For a group of $N$ trajectories, let $Y$ be the ground-truth SID set,
$y_i$ the SID predicted by trajectory $i$, and
$r_i=\mathbf{1}[y_i\in Y]/c_i$, where
$c_i=\sum_{j=1}^{N}\mathbf{1}[y_j=y_i]$.
The total reward within a group equals its number of distinct correct
predictions:
\begin{equation}
    \sum_{i=1}^{N}r_i
    =\left|\{y_i:y_i\in Y,\ 1\leq i\leq N\}\right|.
    \label{eq:coverage-equivalence}
\end{equation}
To see this, group correct rollouts by disease. If a correct SID appears
$c$ times, its total contribution is $c(1/c)=1$; summing over distinct
correct SIDs gives Equation~\ref{eq:coverage-equivalence}. Incorrect
predictions contribute zero regardless of their frequency. The reward
therefore increases when the group covers another correct SID, but not
when it repeats a correct SID or generates a new incorrect one.

Equation~\ref{eq:coverage-equivalence} describes the raw group reward.
GRPO normalizes rewards within each group before updating the policy.
When rewards vary, trajectories predicting less-repeated correct SIDs
receive larger normalized advantages than duplicate or incorrect
trajectories. If all rollouts have the same reward, their normalized
advantages are zero:
an all-identical correct group receives no policy-gradient reinforcement,
while an all-distinct correct group has already attained the maximum
coverage possible with $N$ rollouts. Finite sampling and advantage
normalization do not guarantee a globally uniform policy over all
ground-truth diagnoses.

\subsection{GRPO Optimization Details}
\label{appendix:grpo-optimization}

Negative-advantage tokens use dual clipping. To limit drift, $\widehat{D}_{\mathrm{KL}}$ averages $e^\delta-\delta-1$ over response tokens, with $\delta=\log\pi_{\mathrm{ref}}-\log\pi_\theta$ (numerically clipped).

The original rollouts contribute the GRPO and KL gradients. A second forward pass over answer-replaced branches contributes the multi-positive gradients. These gradients accumulate on the same actor parameters across microbatches, followed by gradient norm clipping and one optimizer step.
At update $t$, eligible rollouts are ordered by rollout index and assigned successive ground-truth SIDs from a fixed enumeration of $Y$, starting at index $((t-1)N)\bmod |Y|$.

\subsection{Catalog-Constrained Beam Search}
\label{appendix:constrained-beam-search}

Let $\mathcal{Q}$ be the catalog of valid three-token SIDs. We build a prefix trie from their tokenizer IDs. For a generated SID prefix $a$, the trie returns the tokens that extend $a$ toward at least one SID in $\mathcal{Q}$:
\begin{equation}
    \mathcal{N}(a)=\{v : a\mathbin{\|}v\text{ is a prefix of some }q\in\mathcal{Q}\}.
\end{equation}
Here $a\mathbin{\|}v$ appends token $v$ to prefix $a$. After a complete SID, only the end-of-sequence token is allowed. At each decoding step, we mask tokens outside $\mathcal{N}(a)$. Starting from the empty prefix, beam search extends each retained path with its allowed next tokens and keeps the $B$ paths with the highest cumulative log probability. This continues through the end-of-sequence token. Completed paths are checked against the catalog and deduplicated before mapping SIDs to diagnoses.

Direct inference uses the checkpoint after reasoning activation. The prompt contains the longitudinal SID history $H$ and disables explicit thinking. The SID-token contribution to the score of diagnosis $d$ is given in Equation~\ref{eq:direct-inference}.
The implementation ranks completed beams by their full sequence scores, which also include the terminating end-of-sequence token. The thinking setting uses the checkpoint after coverage-reward RL with multi-positive supervision. We first sample $R$ reasoning chains. Each chain is appended to the same history prompt, and the trie constrains only the subsequent SID tokens. The resulting local rankings are fused as in Equation~\ref{eq:rrf}.

\section{Additional Experimental Details}
\label{appendix:experimental-details}

\subsection{Dataset Preprocessing}
\label{appendix:dataset-preprocessing}

We process MIMIC-III and MIMIC-IV following prior diagnosis-prediction studies~\citep{ma2018kame,chen2024predictive}. For both datasets, we retain patients with at least two visits and arrange each patient's visits chronologically. Given the diagnoses observed in the historical visits, the task is to predict the complete diagnosis set recorded at the next visit. We split each dataset by patient into training, validation, and test sets with a ratio of 7:1:2, ensuring that all visits from the same patient occur in only one split.

Because MIMIC-IV is substantially larger than MIMIC-III, we construct the experimental MIMIC-IV cohort by sampling 10,000 patients from the patient-level partitions while preserving the 7:1:2 ratio: 7,000 patients from the training split, 1,000 from the validation split, and 2,000 from the test split. Table~\ref{tab:dataset-statistics} reports the cohort statistics after the common preprocessing procedure and before this MIMIC-IV subsampling.

\begin{table}[t]
    \centering
    \caption{Dataset statistics after cohort preprocessing and before subsampling MIMIC-IV.}
    \label{tab:dataset-statistics}
    \small
    \begin{tabular}{lrr}
        \hline
        Dataset & MIMIC-III & MIMIC-IV \\
        \hline
        \# of patients & 5,449 & 79,393 \\
        \# of visits & 14,141 & 408,990 \\
        Avg. \# of visits per patient & 2.60 & 5.15 \\
        Max. \# of visits per patient & 29 & 170 \\
        \# of unique diagnoses & 3,874 & 37,917 \\
        \# of CCS codes & 285 & 842 \\
        \hline
    \end{tabular}
\end{table}

\subsection{Baselines}
\label{appendix:baselines}

We compare \method{} with representative baselines spanning conventional neural predictors, language-assisted predictors, and a zero-shot large language model:
\begin{itemize}
    \item \textbf{RETAIN}~\citep{choi2016retain} uses two reverse-time recurrent attention mechanisms to assign visit-level and variable-level importance weights for accurate and interpretable prediction from longitudinal EHRs.
    \item \textbf{KAME}~\citep{ma2018kame} augments a recurrent patient-history encoder with medical knowledge-graph embeddings and a knowledge-attention mechanism over relevant ontology concepts for next-visit diagnosis prediction.
    \item \textbf{StageNet}~\citep{gao2020stagenet} combines a stage-aware LSTM with stage-adaptive convolution to identify latent health-stage transitions and capture progression patterns for risk prediction.
    \item \textbf{CGL}~\citep{lu2021collaborative} jointly learns patient and disease representations from a patient--disease observation graph and a disease ontology graph, while integrating attention-weighted clinical-text features into sequential prediction.
    \item \textbf{TRANS}~\citep{chen2024predictive} represents each longitudinal EHR as a temporal heterogeneous graph and applies a temporal graph transformer that combines temporal edge features with global positional and local structural encodings.
    \item \textbf{BoxLM}~\citep{tan2025boxlm} fuses language-model semantics with ontology- and EHR-derived hierarchies through box embeddings and models evolving patient states for diagnosis prediction using volume-based similarity.
    \item \textbf{HieraMed}~\citep{KE2026104087} encodes LLM-generated medical-code descriptions with Sentence-BERT, aggregates diagnoses, procedures, and medications in visit-centered heterogeneous graphs, and models patient-state evolution across visits with a Liquid Neural Network for next-visit diagnosis prediction.
    \item \textbf{DeepSeek-V4-Flash (zero-shot)} is prompted directly to perform next-visit diagnosis prediction without in-context demonstrations or task-specific parameter updates.
\end{itemize}
For a fair comparison, we set both the disease embedding dimension and the dimension of the aggregated patient representation to 64 dim for all EHR-trained baselines.

\paragraph{DeepSeek-V4-Flash zero-shot evaluation.}
We evaluate DeepSeek-V4-Flash on the test splits of MIMIC-III and MIMIC-IV without in-context examples or task-specific training. For MIMIC-IV, we use the 10,000-patient cohort. The prompt presents a patient's prior diagnosis codes in chronological visit order and asks for 40 unique next-visit diagnoses ranked by likelihood. It includes neither diagnosis names nor target diagnoses. Non-thinking and thinking use the same prompt and differ only in whether the API's thinking mode is enabled. Both modes use temperature 0 and a maximum generation budget of 8,192 tokens on each dataset.

We evaluate the final answer rather than the model's reasoning text. Predicted codes are normalized, matched to the dataset's diagnosis catalog, and deduplicated while preserving their order. Invalid codes are discarded, and at most 40 predictions are retained. R@$k$ averages the fraction of ground-truth diagnoses retrieved among the top $k$ predictions across visits, for $k\in\{10,20,30,40\}$. 

\subsection{Implementation Details}
\label{appendix:implementation-details}

We use Qwen3-Embedding-8B~\citep{qwen3embedding} to obtain the continuous semantic representation $x_d$ of each disease from its ontology-enriched textual metadata for SID construction. We use Qwen3-1.7B~\citep{yang2025qwen3} as the backbone and optimize all model parameters in each training stage. DeepSeek-V4-Flash is used to generate the disease-level and longitudinal corpora for enriched SID--language alignment and reasoning activation. In the multi-task alignment stage, SID and visit-boundary tokens are appended to the tokenizer vocabulary and their embeddings are trained jointly with the backbone. We train with AdamW, an effective batch size of 1024, a per-GPU micro-batch size of 8, a learning rate of $3\times10^{-4}$, and a maximum sequence length of 1024. Training runs for at most five epochs with early stopping on validation loss. We use 20 warm-up steps followed by cosine learning-rate decay to 10\% of the initial rate.

The reasoning-activation stage initializes from the multi-task alignment checkpoint and retains the same effective batch size, micro-batch size, sequence length, scheduler, and maximum number of epochs, while reducing the learning rate to $1\times10^{-5}$. 

For reinforcement learning, we implement GRPO~\citep{shao2024deepseekmath} with the \textsc{verl} framework. We use a training batch size and PPO mini-batch size of 256, a per-GPU PPO micro-batch size of 8, and a learning rate of $1\times10^{-6}$. For each input, we sample 16 rollouts with temperature 1.0 and top-$p$ 1.0; both the maximum prompt length and maximum response length are 1024. We set the KL-loss coefficient to $1\times10^{-3}$, the entropy coefficient to zero, and the multi-positive loss coefficient to $\lambda_{\mathrm{MP}}=5\times10^{-3}$. The actor is trained for at most ten epochs with gradient checkpointing. All stages are distributed over 4 NVIDIA A800 GPUs.

\subsection{Semantic Structure of Medical SIDs}
\label{app:sid-semantics}

\subsubsection{Randomly Sampled First-Token Cases}
\begin{figure}[H]
    \centering
    \includegraphics[width=0.72\linewidth]{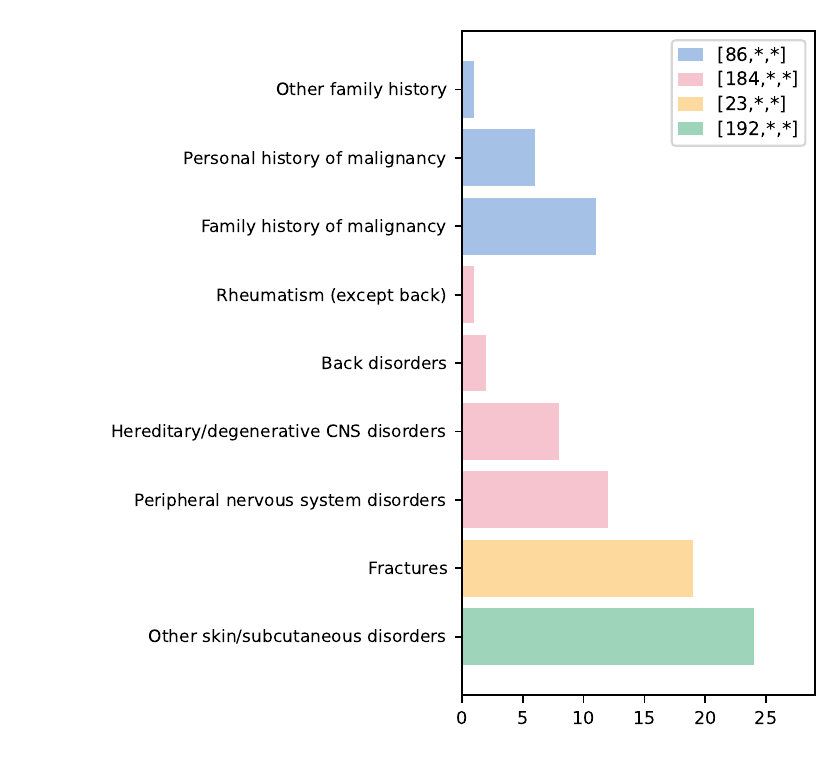}
    \caption{ICD ontology subgroup counts for four randomly sampled first SID
    tokens. Subgroups are the second nodes below the ontology root.}
    \label{fig:sid-prefix-random}
\end{figure}

We sampled four distinct first SID tokens uniformly from the 256 used by the
name plus ICD path variant, using a fixed seed of 2024. Figure~\ref{fig:sid-prefix-random}
shows how their diseases are distributed across ICD ontology subgroups. Each
count represents one distinct disease.

The sampled prefixes have recognizable clinical themes. Of the 18 diseases
beginning with $\langle a_{86}\rangle$, 17 describe a personal or family history
of malignancy. The $\langle a_{184}\rangle$ prefix contains 20 nervous system
diagnoses among 23 diseases, mainly peripheral and hereditary or degenerative
disorders. All 19 diseases under $\langle a_{23}\rangle$ are spinal cord injuries,
and all 24 under $\langle a_{192}\rangle$ are chronic skin ulcers. The three
musculoskeletal diagnoses under $\langle a_{184}\rangle$ also illustrate that a
first token can connect related conditions across ontology branches.

\subsubsection{Semantic Organization Across the Disease Vocabulary}
We measured how well a shared first token groups the 4,491 distinct diseases
by their top-level ICD ontology group and single-level CCS category. The
vocabulary spans 52 top-level groups and 264 CCS categories. For each
token, majority purity is the fraction of its diseases in the most common
category. Weighted purity averages this fraction over diseases. Pair
agreement is the probability that two diseases with the same first token
also have the same category. Both SID variants use the same disease labels.

\begin{table}[H]
    \centering
    \caption{Semantic organization of first SID tokens over 4,491 distinct
    diseases. ICD groups are the first nodes below the ontology root.}
    \label{tab:sid-semantic-organization}
    \small
    \renewcommand{\arraystretch}{1.08}
    \setlength{\tabcolsep}{7pt}
    \begin{tabular}{lcc}
        \hline
        Metric & Name only & Name + ICD path \\
        \hline
        First-token groups pure by ICD / 256 & 58 & \textbf{143} \\
        First-token groups $\geq 80\%$ pure by ICD / 256 & 108 & \textbf{212} \\
        Top-level ICD weighted purity (\%) & 69.36 & \textbf{90.51} \\
        Same ICD group given shared token (\%) & 57.60 & \textbf{86.49} \\
        First-token groups pure by CCS / 256 & 13 & \textbf{44} \\
        First-token groups $\geq 80\%$ pure by CCS / 256 & 36 & \textbf{103} \\
        CCS weighted purity (\%) & 47.63 & \textbf{69.78} \\
        Same CCS category given shared token (\%) & 32.74 & \textbf{57.39} \\
        \hline
    \end{tabular}
\end{table}

Table~\ref{tab:sid-semantic-organization} shows stronger semantic organization
for SIDs learned from names plus ICD paths. The share of diseases assigned to
their token's dominant top-level group rises from 69.36\% to 90.51\%, and 143 rather than
58 first-token groups contain a single top-level group. The same pattern
holds at the finer CCS level: diseases sharing a first token have the same
CCS category with probability 57.39\% for names plus paths, compared with
32.74\% for names alone. These results extend the sampled cases in
Figure~\ref{fig:sid-prefix-random} to the full disease vocabulary.

\subsection{General Ability Evaluation}
\label{appendix:general-ability}

Table~\ref{tab:general-ability} reports general-ability scores for the base
Qwen3-1.7B model and two intermediate training stages. Enriched alignment (S2)
reduces performance on both benchmarks. Adding general reasoning data (S3)
partially recovers these abilities.

\begin{table}[H]
    \centering
    \caption{General-ability evaluation across training stages. Higher scores
    are better.}
    \label{tab:general-ability}
    \small
    \begin{tabular}{lcc}
        \hline
        Model & IFEval & GSM8K \\
        \hline
        Base Qwen3-1.7B & 0.6820 & 0.8832 \\
        S2 = S1 + Enriched Alignment & 0.1460 & 0.5238 \\
        S3 = S2 + General Reasoning & 0.4713 & 0.6300 \\
        \hline
    \end{tabular}
\end{table}

\section{Broader Impacts, Limitations, and Future Work}
\label{appendix:broader-impacts-limitations-future-work}

\subsection{Broader Impacts}
\method{} turns longitudinal diagnosis histories into ranked candidate
diagnoses. In clinical review, these candidates and the thinking mode's
readable traces could help physicians revisit earlier visits and consider
conditions that merit further assessment. Clinical decisions would still rest
on examination, testing, and professional judgment. Prospective studies could
examine how this assistance changes case review and diagnostic decisions.

The compositional medical tokenizer also suggests a generative interface for
large medical vocabularies. Because SID construction uses concept descriptions
and ontology paths, the pipeline could be adapted to other coding systems.
Adding medication, procedure, and laboratory events to the history could allow
diagnosis prediction to use a fuller record. The same representation and
training strategy could then be studied for other EHR prediction tasks.

\subsection{Limitations}
The thinking mode suggests a mechanism for improved diagnosis coverage. A
reasoning chain $\tau$ generated from the SID history $H$ may foreground a
temporal pattern or comorbidity and alter the conditional ranking of a
diagnosis SID $y=q(d)$. Because $\tau$ derives from $H$, it can organize
evidence already in the history and guide the model toward complementary
diagnoses. The gains in Table~\ref{tab:main-results} are consistent with this
account. 

The clinical meaning of a generated trace requires separate study. Plausible
chains of thought can omit factors that influence a prediction, and their
faithfulness varies across models and tasks
\citep{turpin2023language,lanham2023measuring}. The teacher receives a
reference diagnosis when constructing training traces, while the student sees
only $H$ at inference. The clinical accuracy of both kinds of trace and any
artifacts linked to target guidance merit closer study. 

\subsection{Future Work}
To examine the role of the reasoning chain, a controlled analysis could hold
the checkpoint $\theta$, SID history $H$, and answer format fixed and compare
$p_\theta(y\mid H,\tau)$ with $p_\theta(y\mid H)$ for each valid SID
$y=q(d)$, before rank fusion. Changes in SID log scores and ranks would indicate
whether a trace promotes correct diagnoses in the target set $Y$ that direct
decoding ranks lower. Across sampled chains, we could measure how many
additional members of $Y$ enter the top-$k$ candidate lists. Comparing direct and thinking
inference with matched sample counts and computation would help quantify the
contributions of trace conditioning, repeated sampling, and rank fusion.

Clinicians could assess teacher and student traces for factual accuracy,
relevance to the recorded history, and usefulness in reviewing candidate
diagnoses. Changes to selected evidence in $H$ or $\tau$ could test whether
predictions respond to the clinical factors expressed in the trace. We also
plan to study SID vocabularies across coding systems and extend the patient
history to medication, procedure, and laboratory events. These extensions
could support medication recommendation and broader EHR event recommendation.

\end{document}